\documentclass[11pt]{article}
\PassOptionsToPackage{table}{xcolor}
\PassOptionsToPackage{hyperfootnotes=false}{hyperref}
\usepackage[preprint]{acl}
\usepackage{times}
\usepackage{latexsym}
\usepackage[T1]{fontenc}
\usepackage[utf8]{inputenc}
\usepackage{microtype}
\usepackage{inconsolata}
\usepackage{graphicx}
\usepackage{booktabs}
\usepackage{multirow}
\usepackage{makecell}
\usepackage{array}
\usepackage{pifont}
\usepackage{siunitx}
\usepackage{subcaption}
\usepackage{tikz}
\usepackage{amsmath}
\usepackage[most]{tcolorbox}
\usepackage{placeins}
\definecolor{groupblue}{RGB}{214,226,234}
\definecolor{autocolor}{RGB}{255,155,155}
\definecolor{manualcolor}{RGB}{255,140,0}
\definecolor{bestcell}{RGB}{230,245,230}
\newcommand{\cmark}{\ding{51}}
\newcommand{\xmark}{\ding{55}}
\DeclareRobustCommand{\annA}{\tikz[baseline=(char.base)]{\node[shape=rectangle,rounded corners=2pt,fill=autocolor,minimum width=10pt,minimum height=10pt,inner sep=0pt] (char) {\textcolor{black}{\small\textbf{A}}};}}
\DeclareRobustCommand{\annM}{\tikz[baseline=(char.base)]{\node[shape=rectangle,rounded corners=2pt,fill=manualcolor,minimum width=10pt,minimum height=10pt,inner sep=0pt] (char) {\textcolor{black}{\small\textbf{M}}};}}

\title{TriGeoBench: A Multimodal Mathematical Geometry Benchmark with Figures and \texttt{TikZ}}
\author{
  Jing Wang\textsuperscript{1,}\thanks{Equal contribution.}\quad
  Yitong Yang\textsuperscript{1,*}\quad
  Hao Dou\textsuperscript{1,*}\quad
  Yaomin Wu\textsuperscript{1,*} \\
  \textbf{Shuting He\textsuperscript{1}\quad
  Tingfei Shen\textsuperscript{2}\quad
  Jia Wu\textsuperscript{2}\quad
  Jiaqi Li\textsuperscript{1}\quad
  Yinglin Wang\textsuperscript{1,}\thanks{Corresponding author: \texttt{wang.yinglin@shufe.edu.cn}.}} \\
  \normalfont\normalsize\textsuperscript{1}School of Computing and Artificial Intelligence \\
  \normalfont\normalsize\textsuperscript{2}School of Information Management and Engineering \\
  \normalfont\normalsize Shanghai University of Finance and Economics, Shanghai, China \\
  \normalfont\normalsize\texttt{wang.yinglin@shufe.edu.cn}
}
\date{}
\hypersetup{
  pdftitle={TriGeoBench: A Multimodal Mathematical Geometry Benchmark with Figures and TikZ},
  pdfauthor={Jing Wang, Yitong Yang, Hao Dou, Yaomin Wu, Shuting He, Tingfei Shen, Jia Wu, Jiaqi Li, Yinglin Wang},
  pdfsubject={arXiv preprint},
  pdfkeywords={TriGeoBench, multimodal mathematical reasoning, geometry, TikZ, benchmark}
}

\begin{document}
\maketitle

\begin{abstract}
Multimodal Large Language Models show strong vision--language capabilities, yet their geometry-centric reasoning remains insufficiently understood. Existing mathematical benchmarks typically rely on rasterized figures and cover limited geometric knowledge areas and response formats, making it difficult to assess whether models reason over geometric structures or exploit superficial visual patterns. We introduce \textbf{TriGeoBench}, a geometry-centric multimodal benchmark constructed from Chinese middle- and high-school mathematics examinations. TriGeoBench contains 7,000 question--answer pairs and 12,235 geometric figures with \texttt{TikZ} code, spanning plane geometry, solid geometry, analytic geometry, function, and vector geometry across diverse response formats. Each instance forms a \{question--answer pair, image, \texttt{TikZ}\} triplet: question--answer pairs are stored in \LaTeX{}, images as RGB figures, and \texttt{TikZ} preserves geometric topology, symbolic annotations, and relational structure. We evaluate representative open- and closed-source MLLMs through image-to-\texttt{TikZ} and multimodal reasoning probes. Results reveal persistent fine-grained structure and image--text alignment challenges. Oracle experiments with human-curated reference \texttt{TikZ} suggest that accurate executable representations provide complementary geometric cues; they do not evaluate an end-to-end conversion pipeline. TriGeoBench offers a testbed for structure-aware mathematical reasoning.
\end{abstract}
\section{Introduction}
\label{sec:intro}

Multimodal Large Language Models (MLLMs)~\cite{liu2023visual,zhu2023minigpt,hurst2024gpt,team2023gemini,li2023blip} demonstrate strong vision--language capabilities, including visual perception~\cite{liu2024mmbench,hudson2019gqa}, chart understanding~\cite{masry2022chartqa,yang2024askchart}, document comprehension~\cite{mathew2021docvqa,tang2025mtvqa}, visual question answering~\cite{lu2022learn}, and multimodal reasoning~\cite{yue2025mmmu}. Joint visual--textual modeling enables instruction alignment, object and relation recognition, text and layout interpretation, and cross-modal reasoning, supported by large-scale vision--language data.

\begin{figure}[t]
    \centering
    \includegraphics[width=\linewidth]{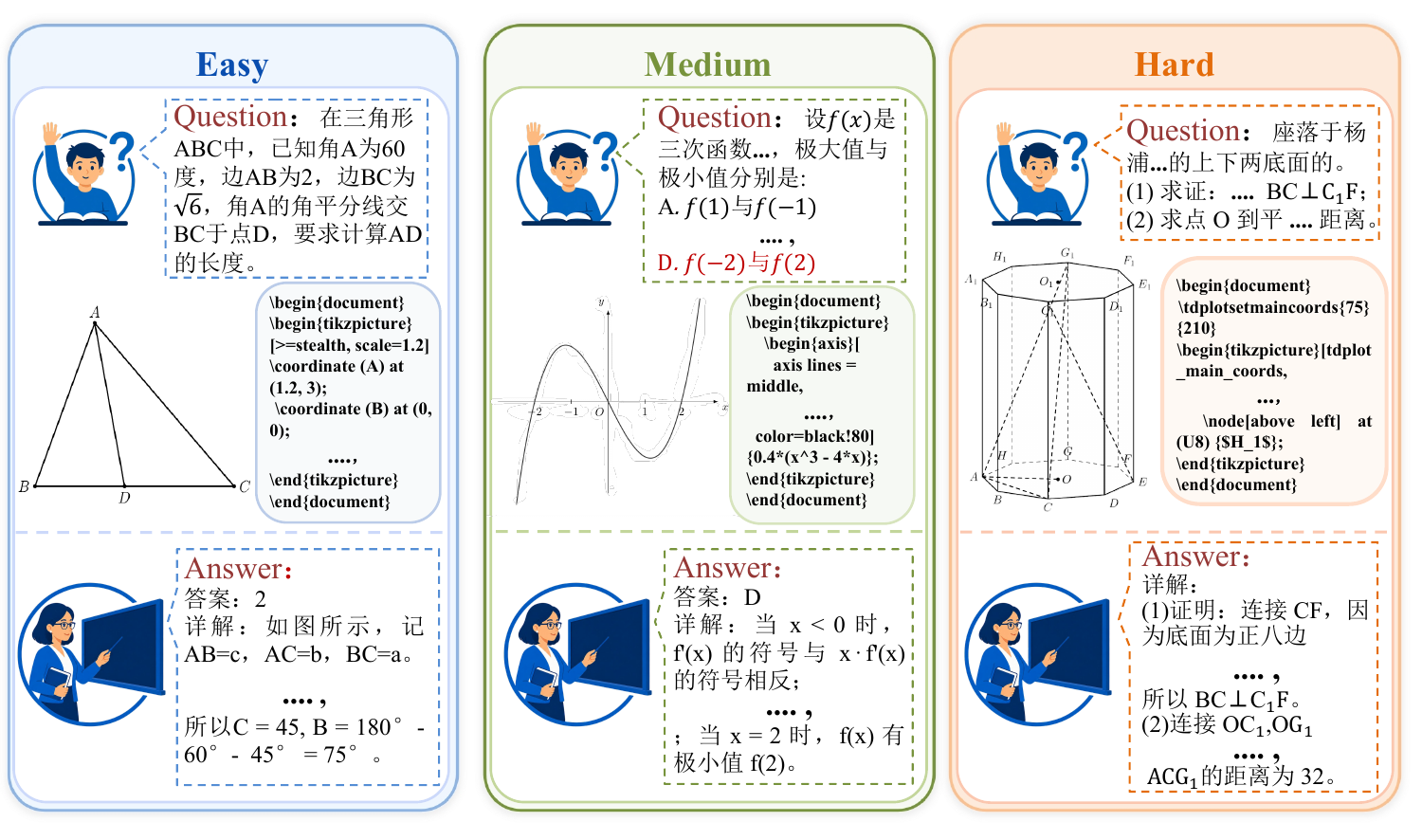}
    \caption{TriGeoBench examples at Easy, Medium, and Hard difficulty levels, with aligned problem text, figures, editable \texttt{TikZ} code, and solutions. Appendix~\ref{app:translations} provides English translations and romanizations.}
    \label{fig:geo_example}
\end{figure}

Multimodal benchmarks evaluate visual recognition, reasoning, domain-specific understanding, hallucination detection, and instruction following~\cite{li2024survey,caffagni2024revolution}. Mathematical reasoning demands understanding mathematical language, interpreting figures, and integrating both to derive answers. Recent benchmarks~\cite{wang2024measuring,10.1007/978-3-031-73242-3_10,li2026visiomath,wang2025solidgeo} evaluate models' integrated understanding and reasoning on problems involving text, symbols, and visual figures. Unlike text-only evaluation~\cite{cobbe2021training,hendrycks2021measuring,gao2024omni}, these benchmarks examine diagram interpretation, symbol--figure alignment, and the integration of multiple information sources for mathematical derivation. However, geometry-centric evaluation retains two limitations: (1) \textbf{Lack of precise structural representations of figures.} Most benchmarks rely on rasterized figures that do not explicitly preserve topology, geometric relations, or symbolic annotations. Consequently, evaluation emphasizes visual appearance and cannot clearly determine whether models reason over precise structures or correctly map mathematical symbols to diagrammatic entities.
(2) \textbf{Limited coverage of geometric knowledge areas and problem formats.} Existing datasets often cover only subsets of geometry and restricted question types, with limited plane and solid geometry or response-format diversity. They therefore provide a coarse view of reasoning.

To address these limitations, we introduce \textbf{TriGeoBench}, a geometry-centric multimodal benchmark for mathematical reasoning and structure-aware diagram understanding. It is constructed from public Chinese middle- and high-school examination problems and solution materials that reflect authentic pedagogical distributions. Manual transcription, figure authoring, and independent human verification yield 7,000 question--answer pairs and 12,235 high-quality figures with corresponding \texttt{TikZ} code. The dataset spans plane geometry, solid geometry, analytic geometry, function, and vector geometry, and includes single-choice, fill-in-the-blank, proof, and essay questions. Each statement is represented in \LaTeX{} and each figure as editable \texttt{TikZ}, preserving topology, symbolic annotations, and relational structure more faithfully than raster images alone. Figure~\ref{fig:geo_example} presents representative examples across difficulty levels.

We evaluate TriGeoBench through unified image-to-\texttt{TikZ} and multimodal reasoning protocols for representative general-purpose MLLMs. Current models still struggle to recover fine-grained relations, annotations, and spatial constraints from figures. For reasoning, we compare text, raster figures, human-curated reference \texttt{TikZ}, and their combination. The reference-code condition is an oracle diagnostic of whether an accurate executable representation is useful once available, rather than an image-to-\texttt{TikZ}-to-reasoning system. Its descriptive gains suggest complementary geometric cues, while raw code can expose exact construction parameters. Our contributions are:
\begin{itemize}
    \item \textbf{TriGeoBench Benchmark.} We introduce a large-scale geometry-centric multimodal benchmark from authentic Chinese examination problems, covering diverse knowledge areas and response formats.
    \item \textbf{Unified Representation.} Each instance is a \{question--answer pair, image, \texttt{TikZ}\} triplet: text is stored in \LaTeX{}, the figure as RGB, and editable \texttt{TikZ} preserves geometric structure and annotations.
    \item \textbf{Broad Evaluation of General-Purpose MLLMs.} Unified probes of image-to-\texttt{TikZ} and mathematical reasoning reveal limitations in precise image--text alignment and structured figure understanding.
\end{itemize}
\section{Related Work}
\label{sec:related}

\textbf{Mathematical and geometry benchmarks.}
Mathematical benchmarks progress from text-only reasoning in MATH~\cite{NEURIPS_DATASETS_AND_BENCHMARKS2021_be83ab3e} to visual evaluation in MathVista~\cite{lu2024mathvista}, MATH-Vision~\cite{wang2024measuring}, and MathVerse~\cite{10.1007/978-3-031-73242-3_10}. Specialized benchmarks VisioMath~\cite{li2026visiomath}, GeoEval~\cite{zhang-etal-2024-geoeval}, and SOLIDGEO~\cite{wang2025solidgeo} show persistent difficulties in fine-grained diagram understanding, image--text alignment, and spatial reasoning. Geometry3K~\cite{lu2021inter} uses formal language to describe texts and diagrams; GeoQA~\cite{chen-etal-2021-geoqa}, GeoQA+~\cite{DBLP:conf/coling/CaoX22}, and UniGeo~\cite{chen-etal-2022-unigeo} introduce programs or proof annotations; PGPS9K~\cite{ijcai2023p0376} further annotates fine-grained geometric primitives and diagram clauses. These datasets use explicit structure but mostly retain raster diagrams or task-specific annotations rather than editable, compilable representations.

\textbf{Graphics program synthesis.}
Symbolic formats such as \LaTeX{}, SVG, and \texttt{TikZ} are readable, editable, and executable. Image-to-\LaTeX{} emphasizes formula and handwriting recognition~\cite{10.1145/3711896.3737436}; DeepSVG~\cite{NEURIPS2020_bcf9d6bd} and VGBench~\cite{zou-etal-2024-vgbench} study vector understanding. DeTikZify~\cite{belouadi2024detikzify}, GeoGramBench~\cite{luo2025geogrambench}, symbolic graphics programming~\cite{chen2025symbolic}, OmniSVG~\cite{yang2025omnisvg}, and InternSVG~\cite{wang2025internsvg} emphasize graphics reconstruction rather than reasoning. Table~\ref{tab:benchmark_comparison} also compares DaTikZ v1/v3 and SciTikZ. TriGeoBench's triplets connect graphics generation with structure-aware reasoning.
\section{TriGeoBench}
\label{sec:method}
\begin{table*}[t]
\centering
{\footnotesize
\setlength{\tabcolsep}{1.8pt}
\begin{tabular*}{\textwidth}{@{\extracolsep{\fill}}l c *{11}{c} l@{}}
\toprule
\multirow{2}{*}{\textbf{Benchmark}}
& \multirow{2}{*}{\centering\textbf{Size}}
& \multicolumn{3}{c}{\textbf{Data Type}}
& \multicolumn{2}{c}{\textbf{Task Type}}
& \multicolumn{5}{c}{\textbf{Knowledge}}
& \multirow{2}{*}{\centering\textbf{Ann.}}
& \multirow{2}{*}{\textbf{Source}} \\
\cmidrule(lr){3-5}\cmidrule(lr){6-7}\cmidrule(lr){8-12}
& & \textbf{Text} & \textbf{Img.} & \textbf{TikZ}
& \textbf{QA} & \makecell{\textbf{Img2}\\\textbf{TikZ}}
& \textbf{Fn.} & \textbf{Vec} & \textbf{PG} & \textbf{SG} & \textbf{AG}
& & \\
\midrule
DaTikZ v1 & 117,832 & \xmark & \cmark & \cmark & \xmark & \cmark & \xmark & \xmark & \xmark & \xmark & \xmark & \annA & \TeX{}.SE and arXiv \\
DaTikZ v3 & 456,469 & \xmark & \cmark & \cmark & \xmark & \cmark & \xmark & \xmark & \xmark & \xmark & \xmark & \annA & \TeX{}.SE and arXiv \\
SciTikZ & 230K & \xmark & \cmark & \cmark & \xmark & \cmark & \cmark & \cmark & \cmark & \cmark & \cmark & \annA\,\annM & Scientific diagrams \\
Geometry3K & 3,002 & \cmark & \cmark & \xmark & \cmark & \xmark & \xmark & \xmark & \cmark & \xmark & \xmark & \annM & Textbooks \\
MATH & 12,500 & \cmark & \xmark & \xmark & \cmark & \xmark & \xmark & \xmark & \xmark & \xmark & \xmark & \annM & Competitions \\
UniGeo & 14,541 & \cmark & \cmark & \xmark & \cmark & \xmark & \xmark & \xmark & \cmark & \xmark & \xmark & \annA\,\annM & GeoQA and web \\
PGPS9K & 9,022 / 4,000 & \cmark & \cmark & \xmark & \cmark & \xmark & \xmark & \xmark & \cmark & \xmark & \xmark & \annM & Geometry3K and web \\
MATH-Vision & 3,040 & \cmark & \cmark & \xmark & \cmark & \xmark & \xmark & \xmark & \cmark & \cmark & \cmark & \annM & Competitions \\
MathVerse & 2,612 & \cmark & \cmark & \xmark & \cmark & \xmark & \cmark & \xmark & \cmark & \cmark & \cmark & \annM & Public datasets \\
SolidGeo & 3,113 & \cmark & \cmark & \xmark & \cmark & \xmark & \xmark & \cmark & \xmark & \cmark & \xmark & \annA\,\annM & K12 / competitions \\
VisioMath & 1,800 / 8,070 & \cmark & \cmark & \xmark & \cmark & \xmark & \cmark & \xmark & \cmark & \cmark & \cmark & \annA\,\annM & Chinese exams \\
\rowcolor{cyan!12}
TriGeoBench & \textbf{7,000 / 12,235} & \cmark & \cmark & \cmark & \cmark & \cmark & \cmark & \cmark & \cmark & \cmark & \cmark & \annM & \textbf{Chinese exams} \\
\bottomrule
\end{tabular*}
\caption{Comparison with mathematical-reasoning and image-to-\texttt{TikZ} datasets. DaTikZ v1 and v3 are shown because they are the releases used by the AutomaTikZ and TikZero baselines, respectively; v2 is an intermediate release. \textbf{Ann.}: \annA{} denotes automatic and \annM{} manual annotation. \textbf{Knowledge}: Fn., function; Vec, vector geometry; PG, plane geometry; SG, solid geometry; AG, analytic geometry.}
\label{tab:benchmark_comparison}
}
\end{table*}
TriGeoBench comprises question--answer pairs, associated images, and corresponding \texttt{TikZ} code. It evaluates whether models reason over structured figures rather than superficial textual or visual cues.

\subsection{Data Collection Principles}
TriGeoBench is built around three principles: diversity, figure-grounded reasoning, and structural representation. Here, figure-grounded means that a diagram contains solution-relevant, nondecorative geometric information used by the reference solution or needed to disambiguate relations. It does not imply strict unsolvability from text alone, because redundant wording, learned priors, or prior exposure may support a correct answer without the figure. Each figure is paired with executable \texttt{TikZ} code to preserve topology, annotations, and geometric relations.

We focus on geometry problems from publicly available Chinese middle- and high-school mathematics exams and their corresponding solution materials from the past three years. These sources reflect authentic pedagogical distributions, enforce standardized mathematical notation, and exhibit naturally coupled text--figure reasoning patterns. As a result, the dataset is well suited not only for evaluating multimodal mathematical reasoning, but also for supporting auxiliary tasks such as image-to-\texttt{TikZ}. Tab.~\ref{tab:benchmark_comparison} presents a detailed comparison between TriGeoBench and existing benchmarks.

\subsection{Annotation}
\label{sec:annotation}

TriGeoBench is constructed through a three-stage annotation pipeline: source collection and preprocessing, manual conversion to symbolic representations, and independent human verification.

\textbf{Collection and preprocessing.}
We collect publicly available Chinese mathematics examination papers from online educational platforms, school repositories, and open exam archives, covering both junior and senior secondary education. These materials typically contain question statements, answer keys, and detailed solution analyses. We extract figure-containing problems related to our target knowledge areas and segment each candidate into structured components, including the question stem, answer, solution, and associated figure(s). To reduce noise from heterogeneous layouts and document formats, we normalize mathematical notation, remove irrelevant document artifacts, and align each question with its corresponding figure region. Annotators filter out problems without valid figures, visually degraded samples, and highly redundant templates, ensuring that retained instances require meaningful text--figure interaction. We also remove irrelevant metadata, personal information, and source-specific formatting artifacts during preprocessing. To reduce leakage and duplication, we perform near-duplicate filtering within the corpus and compare candidates with public geometry and visual-math benchmarks when applicable. Because the sources are public and proprietary pretraining corpora are unknown, these checks reduce obvious overlap but cannot rule out training-set contamination; details are in the appendices.

\textbf{Manual conversion to \LaTeX{} and \texttt{TikZ}.}
To obtain machine-usable and editable representations, annotators manually transcribe each problem statement into \LaTeX{} and author the corresponding \texttt{TikZ} code using the original figure as reference. They follow shared annotation guidelines to preserve mathematical notation, problem semantics, geometric topology, hidden edges, labels, incidence relations, and relative positioning. Complex figures are iteratively compiled and compared with the source diagram until their visual and structural content is faithfully represented.

\textbf{Independent verification and correction.}
Every converted instance undergoes a separate human review against the original source materials. Reviewers check the \LaTeX{} for semantic correctness and the \texttt{TikZ} code for compilability, visual fidelity, and structural consistency with the original figure. In particular, they identify notation errors, missing or distorted problem semantics, text--figure mismatches, and inaccurate geometric topology or label placement. Items with errors are returned for correction, and figures are manually redrawn when needed. This verification stage ensures that the final dataset is syntactically standardized, semantically faithful, and structurally aligned across question text, figures, and \texttt{TikZ} representations. Additional annotation details are provided in the appendices.

The annotation interface presents source images, transcribed \LaTeX{}, and authored \texttt{TikZ} side by side, with compilation checks during editing. The review system records annotation status and acceptance decisions, tracks each instance through correction, and helps avoid duplicated or conflicting edits. Reviewers verify mathematical correctness, text--figure consistency, and rendering quality before accepting an item (Appendix~\ref{app:annotation-details}).

\subsection{Standardized Data Organization}
\begin{figure*}[t]
    \centering
    \includegraphics[width=\textwidth]{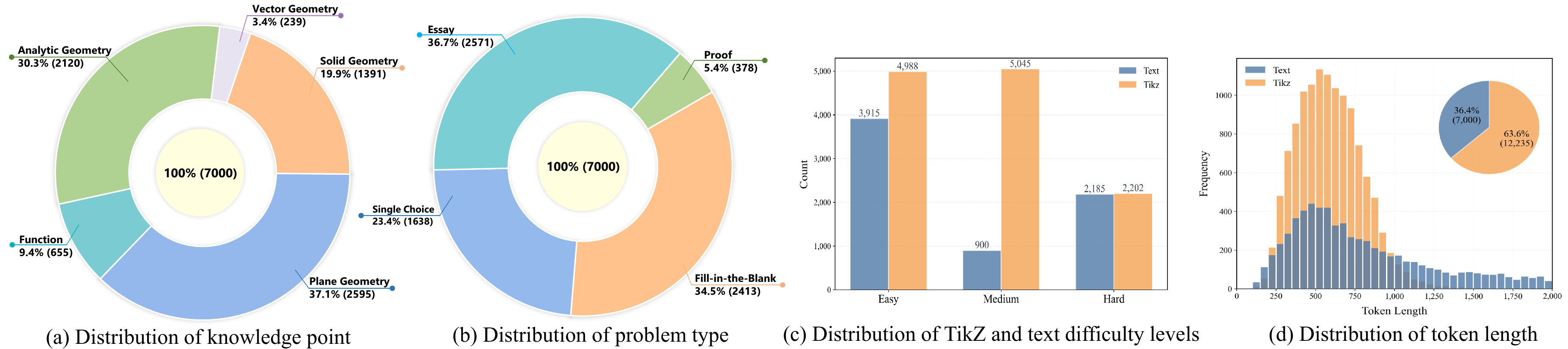}
    \caption{Detailed statistics of the TriGeoBench dataset. In (c) and (d), \textit{Text} denotes the question--answer pair, while \texttt{TikZ} denotes the code corresponding to the image.}
    \label{fig:distribution}
\end{figure*}

To support downstream usage, we decouple the textual and graphical components of each instance. Specifically, the problem content is stored in \LaTeX{} form, while each figure is stored separately in \texttt{TikZ} form. In the text, the original figure locations are replaced with explicit placeholders that specify the figure identity and local position, with each placeholder linked to the corresponding figure through the question ID and figure index. The unified representation enables flexible training and fine-tuning for a range of downstream tasks. To further support fine-grained evaluation and analysis, we categorize TriGeoBench along three dimensions:
\begin{itemize}
    \item \textbf{Knowledge point.} Each problem is labeled with one or more knowledge areas, including plane geometry, solid geometry, analytic geometry, function, and vector geometry. Since many problems involve multiple topics, we use both the question stem and the official solution for model-assisted labeling, and manually resolve ambiguous or overlapping cases. Fig.~\ref{fig:distribution}~(a) shows the distribution across knowledge areas.
    \item \textbf{Problem type.} We classify questions into four response formats: single choice, fill-in-the-blank, proof-based, and essay questions, with their statistical distribution shown in Fig.~\ref{fig:distribution}~(b). Initial labels are proposed by a large language model based on the question form and answer structure, and are subsequently verified by human annotators.
    \item \textbf{Figure complexity and problem difficulty.} We annotate both dimensions in two stages. First, K-means initializes three heuristic strata: reference-solution length approximates problem difficulty, while \texttt{TikZ}-code length approximates figure complexity, as shown in Fig.~\ref{fig:distribution}~(d). Second, experienced teachers and educational experts calibrate the labels. They adjust difficulty according to solution strategy, inference depth, and text--figure interaction, and refine complexity using labeled entities, auxiliary constructions, geometric relations, and spatial dependencies. These strata describe dataset-level difficulty and do not guarantee monotonic accuracy for every model. Fig.~\ref{fig:distribution}~(c) shows the difficulty distribution.
\end{itemize}

After manual conversion and independent human verification, TriGeoBench comprises 7,000 question--answer pairs paired with 12,235 high-quality geometric figures. Compared with prior benchmarks, TriGeoBench offers broader topical coverage, more diverse question formats, and native \texttt{TikZ}-based figure representations, making it suited for both multimodal mathematical reasoning and figure generation tasks.
\section{Experiments}
\label{sec:experiments}

We design two tasks: Image-to-\texttt{TikZ} evaluates conversion into editable code that recovers visual structure, geometric relations, and annotations; Multimodal Mathematical Reasoning assesses joint statement--diagram interpretation, symbol--figure alignment, and answer derivation. Together, they probe diagram generation and geometric reasoning.

\noindent \textbf{Evaluation Models.} We evaluate a broad spectrum of MLLMs on TriGeoBench, including both open-source and closed-source models. The open-source set covers eight model families: Qwen3-VL models~\cite{bai2025qwen3} of different scales, Qwen2.5-VL~\cite{bai2025qwen25}, GLM-4.5V and GLM-4.1V-9B-Thinking~\cite{hong2025glm}, Llama-3.2-Vision~\cite{meta2024llama32hf}, Gemma3~\cite{gemma2025}, LLaVA-OneVision~\cite{li2024llava}, and InternVL3~\cite{zhu2025internvl3}. The closed-source set includes four representative models: Gemini-3.1-Pro~\cite{google2026gemini31pro}, GPT-5.4~\cite{chatgpt54}, Seed-2.0-Pro~\cite{bytedance2025doubao15pro}, and Qwen3.5-Plus~\cite{qwen2026qwen35plus}.

% \noindent \textbf{Evaluation Metrics.} For the Image-to-\texttt{TikZ} task, we use four complementary metrics. \textbf{Gen Acc} measures whether the generated \texttt{TikZ} code can be successfully compiled into an image, evaluating its syntactic correctness and executability. \textbf{Image Metrics} assess the visual consistency between the rendered image from the generated code and the ground-truth image. \textbf{\texttt{TikZ} Metrics} compare the generated code with the ground-truth \texttt{TikZ} to quantify structural and symbolic similarity. \textbf{Human Acc} measures, via human evaluation, whether the rendered image is visually and semantically consistent with the ground-truth image.

\noindent \textbf{Evaluation Metrics.} For the Image-to-\texttt{TikZ} task, we use four complementary metrics. \textbf{Gen Acc} measures whether the generated \texttt{TikZ} code can be successfully compiled, evaluating executability. \textbf{Image Metrics} assess visual consistency between rendered predictions and ground-truth images. \textbf{\texttt{TikZ} Metrics} compare generated and reference code to quantify symbolic and structural similarity. \textbf{Human Acc} measures the proportion of rendered figures judged by humans as visually and mathematically faithful. For mathematical reasoning, we report answer accuracy using an LLM-based grader, counting each sub-question independently. Formal definitions are provided in the appendices.

\noindent \textbf{Evaluation Datasets.} We construct the evaluation set by sampling 1,000 instances from TriGeoBench across different difficulty levels, including 500 Easy, 300 Medium, and 200 Hard problems. This stratified sampling strategy ensures that the test set covers a broad range of reasoning demands and figure complexities. The remaining instances are used as the training set for fine-tuning small-scale multimodal large language models.

\subsection{A Unified Image-to-\texttt{TikZ} Probe}
\begin{figure*}[t]
    \centering
    \begin{subfigure}{0.32\textwidth}
        \centering
        \includegraphics[width=\linewidth]{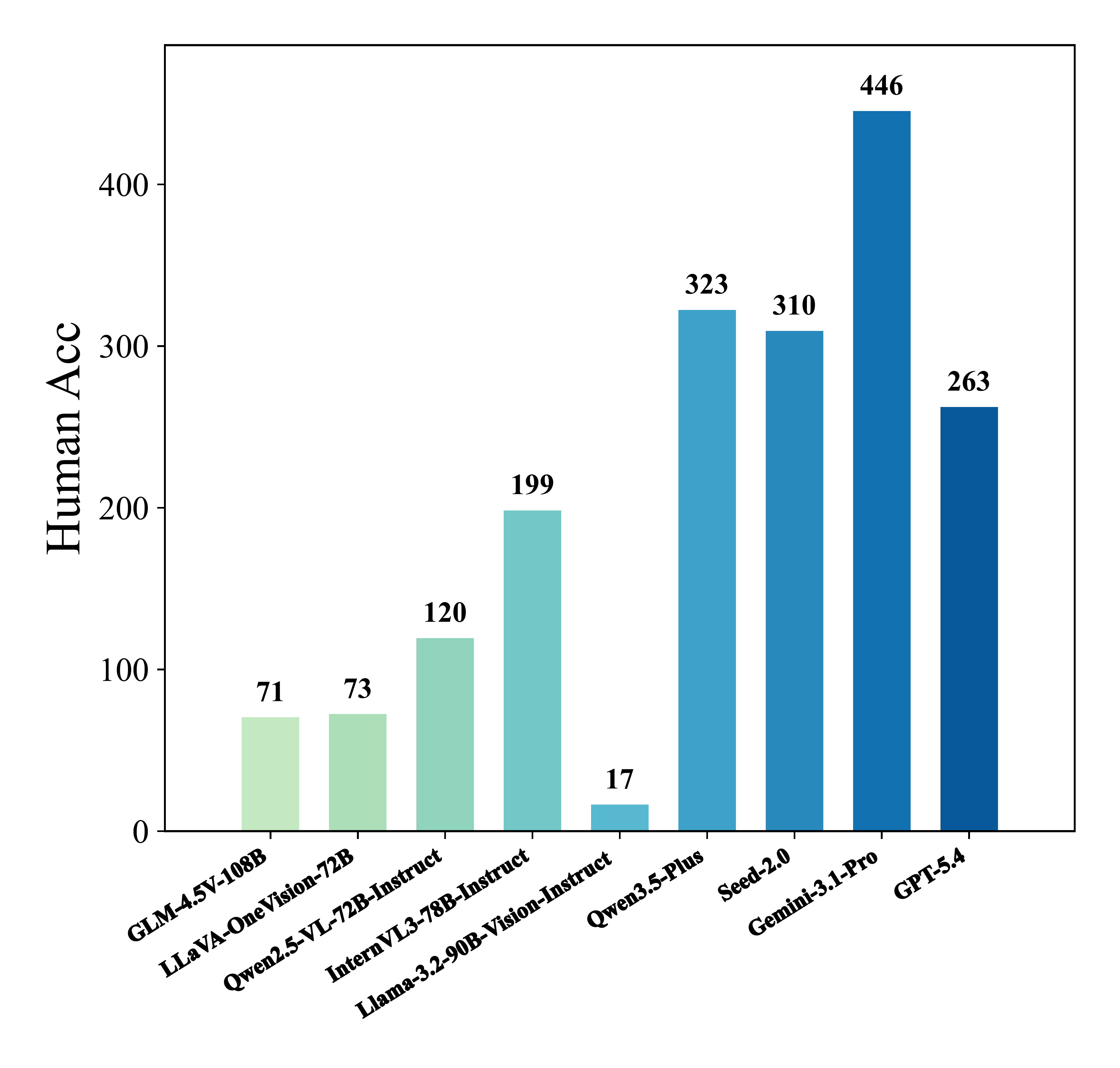}
        \caption{Easy}
    \end{subfigure}
    \hfill
    \begin{subfigure}{0.32\textwidth}
        \centering
        \includegraphics[width=\linewidth]{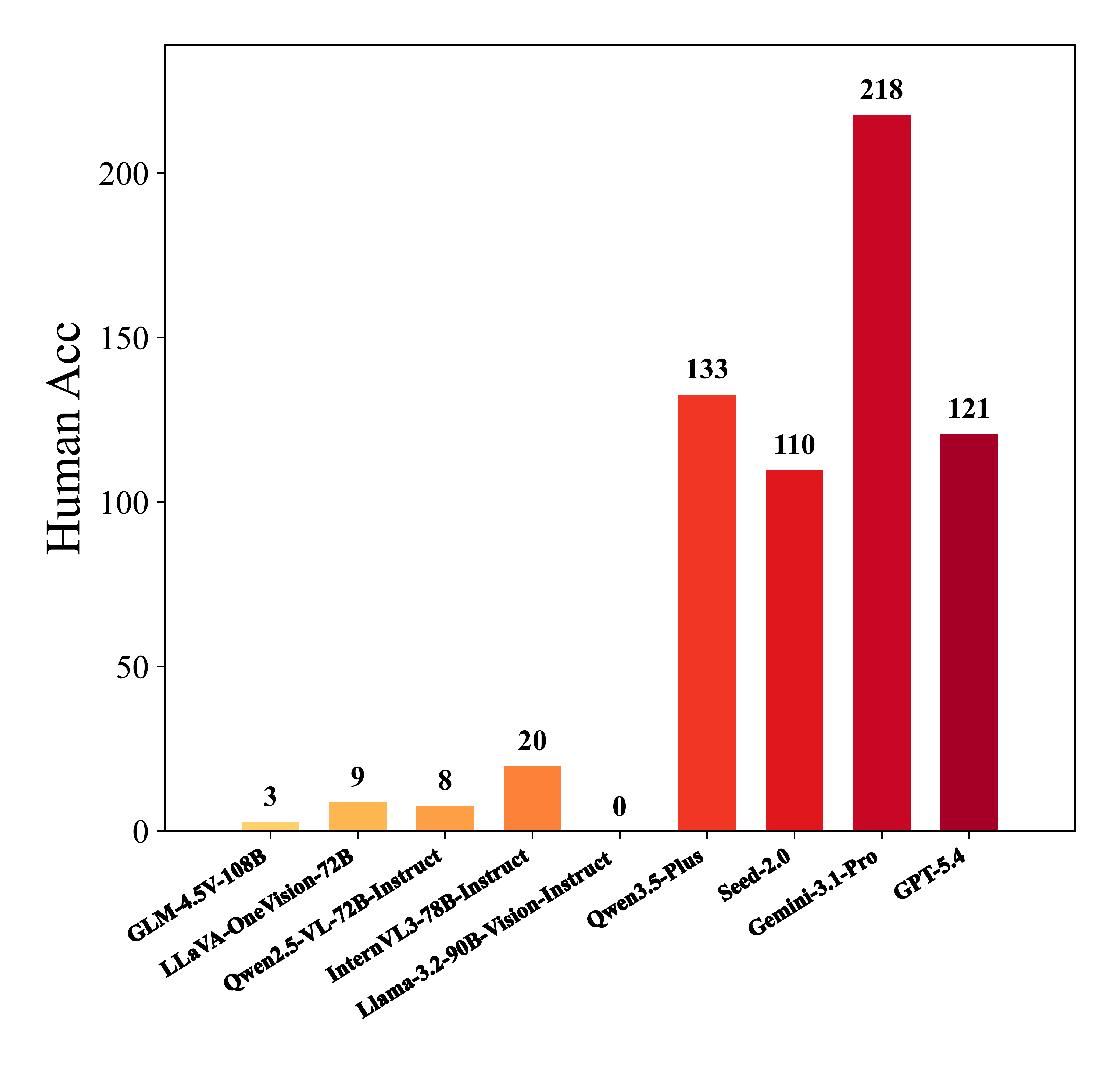}
        \caption{Medium}
    \end{subfigure}
    \hfill
    \begin{subfigure}{0.32\textwidth}
        \centering
        \includegraphics[width=\linewidth]{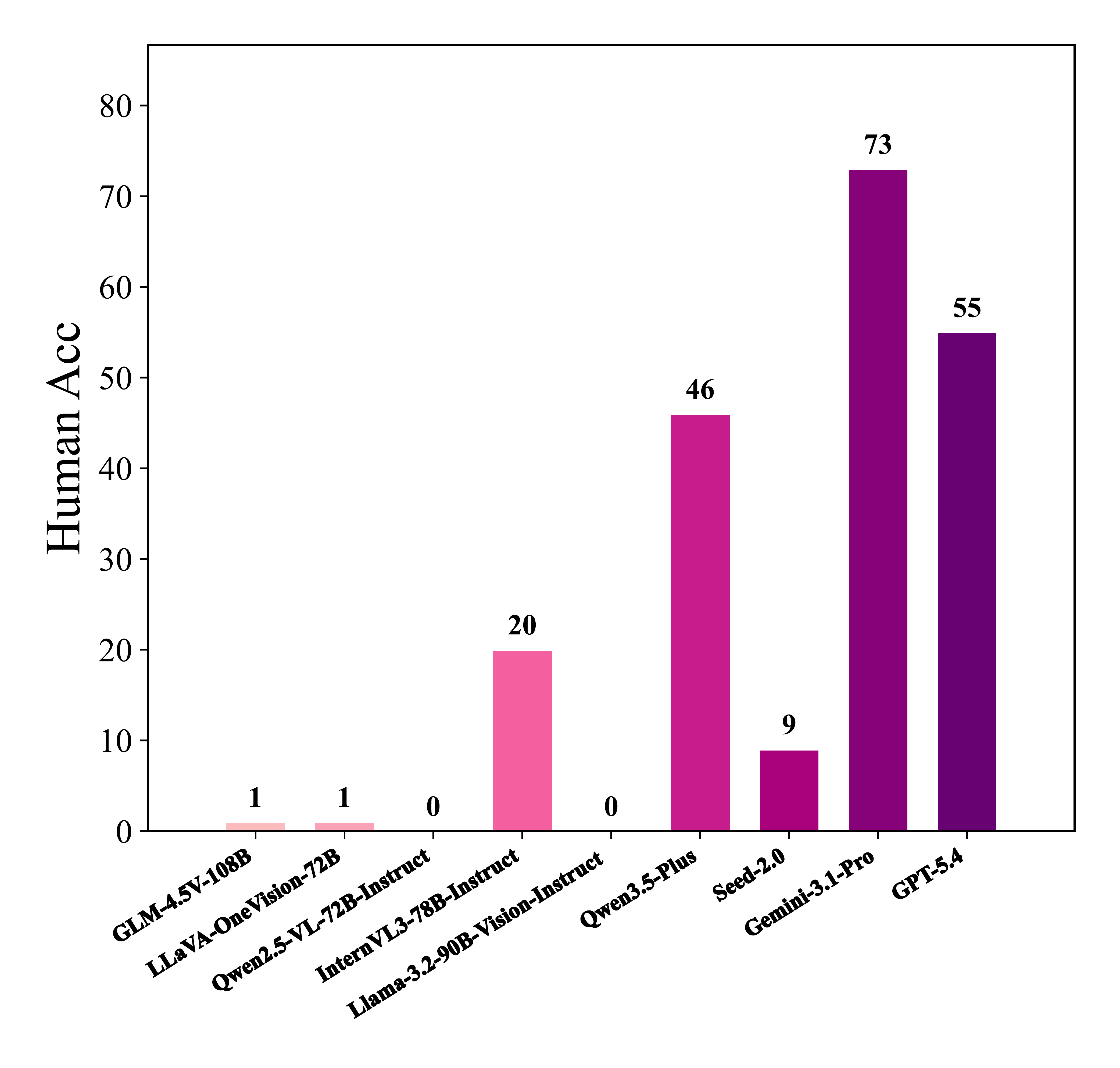}
        \caption{Hard}
    \end{subfigure}
    \caption{Human accuracy of different models on Image-to-\texttt{TikZ} tasks across difficulty levels.}
    \label{fig:viz_i2l}
\end{figure*}
\begin{table*}[t]
\centering
{\footnotesize
\setlength{\tabcolsep}{1.8pt}
\begin{tabular*}{\textwidth}{@{\extracolsep{\fill}}l*{9}{c}@{}}
\toprule
\multirow{2}{*}{\textbf{Model}}
& \multirow{2}{*}{\makecell{\textbf{Gen}\\\textbf{Acc} $\uparrow$}}
& \multicolumn{4}{c}{\textbf{Image Metrics}}
& \multicolumn{3}{c}{\textbf{\texttt{TikZ} Metrics}}
& \multirow{2}{*}{\makecell{\textbf{Human}\\\textbf{Acc} $\uparrow$}} \\
\cmidrule(lr){3-6}\cmidrule(lr){7-9}
& & \makecell{\textbf{SSIM}\\$\times 10^2\!\uparrow$} & \makecell{\textbf{KID}\\$\times 10^3\!\downarrow$} & \makecell{\textbf{CLIP}\\$\times 10^2\!\uparrow$} & \makecell{\textbf{DINO}\\$\times 10^2\!\uparrow$} & \textbf{cBLEU $\uparrow$} & \makecell{\textbf{MTE}\\$\times 10^2\!\uparrow$} & \textbf{TED $\downarrow$} & \\
\midrule
\rowcolor{groupblue}\multicolumn{10}{c}{\textbf{Open-Source Models}} \\
Llama-3.2-90B-Vision-Instruct & 80.5\% & 77.54 & 30.9 & 77.86 & 54.61 & 35.27 & 64.22 & 74.67 & 1.7\% \\
GLM-4.5V-108B & 76.4\% & 75.45 & \bfseries 4.0 & 94.54 & 87.17 & 45.91 & 76.37 & 6.55 & 7.5\% \\
Qwen2.5-VL-72B-Instruct & 93.5\% & 80.66 & 67.8 & 70.47 & 41.60 & 50.39 & 76.52 & 7.26 & 12.8\% \\
LLaVA-OneVision-72B & 68.8\% & \cellcolor{bestcell}\bfseries 89.03 & 96.3 & 81.56 & 66.62 & 22.39 & 50.61 & 56.60 & 4.1\% \\
InternVL3-78B-Instruct & 78.4\% & 74.52 & 6.9 & 92.80 & 82.52 & 39.53 & 70.92 & 9.47 & 8.3\% \\
\midrule
\rowcolor{groupblue}\multicolumn{10}{c}{\textbf{Closed-Source Models}} \\
Qwen3.5-Plus & 88.2\% & 86.33 & 6.7 & \bfseries 97.03 & \bfseries 94.12 & \cellcolor{bestcell}\bfseries 63.91 & \bfseries 82.35 & \bfseries 5.28 & \bfseries 50.2\% \\
Seed-2.0-Pro & 93.0\% & 85.70 & 8.2 & 94.70 & 89.66 & 52.44 & 78.44 & 5.83 & 42.9\% \\
Gemini-3.1-Pro & \cellcolor{bestcell}\textbf{96.1\%} & \bfseries 86.44 & \cellcolor{bestcell}\bfseries 2.2 & \cellcolor{bestcell}\bfseries 97.76 & \cellcolor{bestcell}\bfseries 95.92 & \bfseries 63.78 & \cellcolor{bestcell}\bfseries 82.73 & \cellcolor{bestcell}\bfseries 3.89 & \cellcolor{bestcell}\textbf{73.7\%} \\
GPT-5.4 & \textbf{95.6\%} & 80.72 & 11.7 & 94.35 & 88.20 & 53.08 & 71.53 & 6.57 & 43.9\% \\
\bottomrule
\end{tabular*}
\caption{TriGeoBench results. Green shading marks the best in each column; bold marks the second-best.}
\label{tab:model_comparison}
}
\end{table*}
\begin{table*}[t]
\centering
{\footnotesize
\setlength{\tabcolsep}{0.6pt}
\begin{tabular}{@{}p{79pt}p{47pt}*{9}{>{\centering\arraybackslash}p{35pt}}@{}}
\toprule
\multirow{2}{*}{\textbf{Model}}
& \multirow{2}{*}{\textbf{Stage}}
& \multirow{2}{*}{\makecell{\textbf{Gen}\\\textbf{Acc} $\uparrow$}}
& \multicolumn{4}{c}{\textbf{Image Metrics}}
& \multicolumn{3}{c}{\textbf{\texttt{TikZ} Metrics}}
& \multirow{2}{*}{\makecell{\textbf{Human}\\\textbf{Acc} $\uparrow$}} \\
\cmidrule(lr){4-7}\cmidrule(lr){8-10}
& & & \makecell{\textbf{SSIM}\\$\times 10^2\!\uparrow$} & \makecell{\textbf{KID}\\$\times 10^3\!\downarrow$} & \makecell{\textbf{CLIP}\\$\times 10^2\!\uparrow$} & \makecell{\textbf{DINO}\\$\times 10^2\!\uparrow$} & \textbf{cBLEU $\uparrow$} & \makecell{\textbf{MTE}\\$\times 10^2\!\uparrow$} & \textbf{TED $\downarrow$} & \\
\midrule
\multirow{2}{*}{\textbf{Qwen3-VL-2B}} & Base model & \textbf{77.7\%} & \bfseries 86.95 & 64.0 & 85.82 & 73.26 & 38.34 & 61.99 & 228.24 & 8.8\% \\
& LoRA-SFT & 72.7\% & 77.35 & \bfseries 2.9 & \bfseries 95.53 & \textbf{88.01} & \bfseries 69.39 & \bfseries 85.67 & \bfseries 4.26 & \textbf{12.0\%} \\
\specialrule{0.25pt}{1pt}{1pt}
\multirow{2}{*}{\textbf{Qwen3-VL-4B}} & Base model & \textbf{80.1\%} & 80.63 & 4.4 & 94.32 & 86.53 & 42.58 & 65.88 & 58.52 & 5.4\% \\
& LoRA-SFT & 71.6\% & \bfseries 82.49 & \bfseries 2.5 & \bfseries 95.69 & \textbf{89.25} & \bfseries 69.71 & \bfseries 85.88 & \bfseries 4.11 & \textbf{13.4\%} \\
\specialrule{0.25pt}{1pt}{1pt}
\multirow{2}{*}{\textbf{Qwen3-VL-8B}} & Base model & 60.1\% & 81.26 & 3.8 & \bfseries 95.38 & 88.27 & 40.91 & 64.12 & 75.16 & 0.6\% \\
& LoRA-SFT & \textbf{75.6\%} & \bfseries 82.89 & \bfseries 3.8 & 94.90 & \textbf{88.86} & \bfseries 69.85 & \bfseries 85.86 & \bfseries 4.27 & \cellcolor{bestcell}\textbf{27.0\%} \\
\specialrule{0.25pt}{1pt}{1pt}
\multirow{2}{*}{\textbf{\makecell[l]{GLM-4.1V-9B-\\Thinking}}} & Base model & 74.6\% & 80.72 & 14.5 & 91.18 & 87.73 & \bfseries 64.50 & \bfseries 83.65 & \bfseries 8.28 & 0.6\% \\
& LoRA-SFT & \cellcolor{bestcell}\textbf{95.5\%} & \bfseries 81.66 & \bfseries 3.7 & \bfseries 95.12 & \textbf{88.36} & 39.33 & 70.05 & 11.18 & \textbf{4.2\%} \\
\specialrule{0.25pt}{1pt}{1pt}
\multirow{2}{*}{\textbf{Llama-3.2VL-11B}} & Base model & 28.7\% & \cellcolor{bestcell}\bfseries 89.64 & 139.5 & 75.16 & 57.10 & 8.45 & 22.75 & 139.14 & 0.3\% \\
& LoRA-SFT & \textbf{88.0\%} & 82.40 & \cellcolor{bestcell}\bfseries 1.0 & \cellcolor{bestcell}\bfseries 96.34 & \cellcolor{bestcell}\textbf{91.25} & \cellcolor{bestcell}\bfseries 70.09 & \cellcolor{bestcell}\bfseries 85.88 & \cellcolor{bestcell}\bfseries 4.08 & \textbf{15.0\%} \\
\specialrule{0.25pt}{1pt}{1pt}
\multirow{2}{*}{\textbf{Gemma3-12B}} & Base model & \textbf{87.0\%} & \bfseries 87.81 & 87.1 & 81.60 & 65.34 & 47.20 & 75.57 & 7.26 & 0.6\% \\
& LoRA-SFT & 72.7\% & 82.11 & \bfseries 2.1 & \bfseries 95.84 & \textbf{89.81} & \bfseries 62.93 & \bfseries 80.86 & \bfseries 5.42 & \textbf{11.3\%} \\
\bottomrule
\end{tabular}
\caption{Results for six small-scale MLLMs before and after fine-tuning. Bold marks the better result within a model pair; green shading marks the best result in a column.}
\label{tab:lora_comparison}
}
\end{table*}

\noindent \textbf{Scope and Interpretation.} All problem statements and evaluation prompts are in Chinese, so scores jointly reflect Chinese comprehension, mathematical reasoning, visual understanding, and structured-input use; model comparisons are not language-controlled estimates of geometry ability alone. Text+\texttt{TikZ} uses human-curated reference code and serves as an oracle diagnostic, not an end-to-end conversion pipeline. We report descriptive point estimates from a fixed run and do not treat small differences as statistically conclusive. Public sources and unknown proprietary pretraining corpora also prevent a complete contamination guarantee.

Our goal is to probe general-purpose MLLMs under a unified prompting protocol, not to establish state-of-the-art performance against task-specific systems. DeTikZify, AutomaTikZ, and TikZero use distinct training corpora and task-specific optimization; their direct evaluation on TriGeoBench remains future work.

\noindent \textbf{Evaluation of MLLMs on TriGeoBench.} 
Tab.~\ref{tab:model_comparison} summarizes the performance of representative open-source and closed-source MLLMs on the TriGeoBench evaluation set. 
Closed-source models have higher human-evaluated point estimates than open-source models, indicating limited open-source performance on this Chinese geometry-focused generation probe. 
Among all evaluated models, Gemini-3.1-Pro achieves the best overall performance, with a generation accuracy (Gen Acc) of 96.1\% and a human-evaluated accuracy (Human Acc) of 73.7\%. 
This suggests that the model can not only generate compilable outputs at a high rate, but also more frequently produce geometrically correct results. 
In contrast, although GPT-5.4 also attains a high generation accuracy of 95.6\%, its human-evaluated accuracy is only 43.9\%, revealing a clear gap between compilation success and actual geometric correctness. 
% For open-source models, although some models achieve relatively high scores on certain automatic metrics, their human-evaluated accuracy remains generally low. 
% This indicates that these models can often generate syntactically valid or visually plausible outputs, but still struggle to reliably preserve fine-grained geometric relations, symbolic annotations, and structural constraints. 
% In other words, the visual similarity or code similarity captured by automatic metrics does not always correspond to final geometric correctness, further highlighting the necessity of human evaluation for assessing geometric semantic consistency. 
These results show that automatic metrics alone cannot fully capture geometric correctness, making human evaluation necessary for assessing structural and semantic fidelity.
We further analyze human-evaluated accuracy across different complexity levels for all models, as shown in Fig.~\ref{fig:viz_i2l}. 
Open-source models have particularly low point estimates on Medium and Hard samples, where matching the input geometry remains difficult. 
Closed-source models have higher estimates across the reported complexity strata. 
% However, even closed-source models show a noticeable decline in accuracy as sample complexity increases. 
% This trend suggests that multi-object relations, symbolic constraints, and compositional spatial structures in complex geometric figures remain key bottlenecks for current MLLMs.
Overall Human Acc tends to decrease with figure complexity, although these strata need not produce a monotonic ordering for every model.

\noindent \textbf{Inference and fine-tuning evaluation on small-scale MLLMs.} 
To further validate the effectiveness and discriminative ability of TriGeoBench, we conduct LoRA fine-tuning experiments on several recent representative small-scale vision--language models, with results reported in Tab.~\ref{tab:lora_comparison}. 
For a fair comparison, all models are fine-tuned under the same LoRA configuration, including learning rate, rank, and number of epochs. 
% This setup allows us to examine how different models adapt to image-to-\texttt{TikZ} generation under the same supervision and training budget.
This controlled setup enables a direct comparison of model adaptability under the same training budget.
The results on Gen Acc show that LoRA fine-tuning does not consistently improve all models. 
For instance, Qwen3-VL-2B and Qwen3-VL-4B even exhibit decreased Gen Acc after fine-tuning. 
% We attribute this mainly to the strong formatting and compilability requirements of \texttt{TikZ} generation. 
% Different geometric images may require different packages, environments, or drawing structures, while standard LoRA-SFT only fits reference answers through supervised learning and does not explicitly enforce syntactic completeness, dependency correctness, or compilation validity. 
% Thus, fine-tuned models may generate code that is locally closer to the reference but still fails due to missing packages, mismatched environments, or minor syntax errors. 
This suggests that supervised fine-tuning alone is insufficient for strongly format-constrained tasks, and future work may benefit from compilation-aware training signals, such as format rewards, compilation-success rewards, or image-consistency rewards.

For Image Metrics, fine-tuning also does not yield consistent improvements. 
Unlike natural images, geometric figures often contain large blank regions, simple line segments, and sparse symbols, allowing general-purpose visual metrics such as CLIP and DINO to assign relatively high scores even to imperfect outputs. 
% However, geometric correctness depends on preserving line segments, angles, arcs, annotations, and spatial relations. 
% Since existing image metrics mainly capture global semantic or visual similarity, they may not accurately reflect whether the rendered \texttt{TikZ} image is geometrically correct. 
% Therefore, higher Image Metrics do not necessarily imply better geometric reasoning, and lower scores do not always indicate worse generated figures.
Therefore, image-level similarity metrics are useful but insufficient for evaluating fine-grained geometric faithfulness.
In contrast, TikZ Metrics generally improve more substantially after fine-tuning, indicating that LoRA-SFT helps models learn domain-specific code patterns, command usage, and structured representations. 
Improvements in cBLEU, MTE, and TED suggest that fine-tuned models generate code closer to the reference \texttt{TikZ} programs at the token, edit-distance, or tree-structure level. 
Nevertheless, the same geometric figure can often be represented by multiple valid \texttt{TikZ} programs using different coordinate systems, drawing orders, macros, or equivalent constructions. 
Thus, code-level metrics reflect alignment with the reference coding style, but cannot fully determine whether the generated code produces a correct image. 
% The results of GLM-4.1V-9B-Thinking further illustrate this mismatch, where strong TikZ Metrics do not always align with final generation accuracy.
To address these limitations, we include human evaluation to better assess geometric structures and symbolic relationships. 
Human-evaluated point estimates increase after LoRA fine-tuning for every model, although small differences are not statistically conclusive. 
This pattern suggests that LoRA-SFT can aid adaptation to image-to-\texttt{TikZ} generation despite discrepancies between metrics and generation quality.

\subsection{Multimodal Mathematical Reasoning}

\begin{table*}[t]
\centering
{\footnotesize
\setlength{\tabcolsep}{1.35pt}
\begin{tabular*}{\textwidth}{@{\extracolsep{\fill}}l*{16}{c}@{}}
\toprule
\multirow{2}{*}{\textbf{Model}}
& \multicolumn{4}{c}{\textbf{Text-only}}
& \multicolumn{4}{c}{\textbf{Text+Figure}}
& \multicolumn{4}{c}{\textbf{Text+\texttt{TikZ}}}
& \multicolumn{4}{c}{\textbf{Text+Figure+\texttt{TikZ}}} \\
\cmidrule(lr){2-5}\cmidrule(lr){6-9}\cmidrule(lr){10-13}\cmidrule(lr){14-17}
& \textbf{Easy} & \makecell{\textbf{Med.}} & \textbf{Hard} & \textbf{Avg.}
& \textbf{Easy} & \makecell{\textbf{Med.}} & \textbf{Hard} & \textbf{Avg.}
& \textbf{Easy} & \makecell{\textbf{Med.}} & \textbf{Hard} & \textbf{Avg.}
& \textbf{Easy} & \makecell{\textbf{Med.}} & \textbf{Hard} & \textbf{Avg.} \\
\midrule
\rowcolor{groupblue}\multicolumn{17}{c}{\textbf{Open-Source Models}} \\
Qwen2.5-VL-72B & 33.29 & 28.05 & 28.81 & 30.67 & 37.39 & 28.88 & 28.02 & 32.79 & 38.57 & 29.73 & 28.98 & 33.90 & 41.10 & 30.98 & 22.99 & 34.94 \\
GLM-4.5V-108B & 49.31 & 38.17 & 34.81 & 43.13 & 48.96 & 38.37 & 39.30 & 43.47 & 51.05 & 39.51 & 29.67 & 43.79 & 52.14 & 40.06 & 33.15 & 44.98 \\
Llama-3.2-90B-Vision & 13.04 & 14.57 & 10.38 & 13.33 & 12.78 & 14.67 & 10.64 & 13.26 & 14.57 & 13.42 & 16.39 & 14.32 & 16.88 & 14.24 & 12.97 & 15.34 \\
LLaVA-OneVision-72B & 21.88 & 17.11 & 16.09 & 18.95 & 22.25 & 19.19 & 17.61 & 20.03 & 21.37 & 22.46 & 16.58 & 19.60 & 24.29 & 14.77 & 17.87 & 20.59 \\
InternVL3-78B & 30.39 & 21.67 & 23.45 & 26.59 & 29.28 & 22.58 & 24.79 & 26.64 & 29.69 & 26.90 & 23.59 & 26.88 & 31.06 & 22.22 & 25.34 & 27.68 \\
\midrule
\rowcolor{groupblue}\multicolumn{17}{c}{\textbf{Closed-Source Models}} \\
Qwen3.5-Plus & \bfseries 73.87 & \bfseries 80.13 & \bfseries 68.57 & \bfseries 75.81 & \bfseries 73.67 & \bfseries 77.69 & \bfseries 64.44 & \bfseries 74.20 & \bfseries 75.65 & \bfseries 80.49 & \bfseries 73.63 & \bfseries 77.36 & \bfseries 75.37 & \bfseries 76.86 & \bfseries 65.57 & \bfseries 74.76 \\
Seed-2.0-Pro & \cellcolor{bestcell}\bfseries 79.10 & \cellcolor{bestcell}\bfseries 84.63 & \cellcolor{bestcell}\bfseries 82.70 & \cellcolor{bestcell}\bfseries 81.76 & \cellcolor{bestcell}\bfseries 83.29 & \cellcolor{bestcell}\bfseries 84.44 & \cellcolor{bestcell}\bfseries 85.56 & \cellcolor{bestcell}\bfseries 84.02 & \cellcolor{bestcell}\bfseries 84.88 & \cellcolor{bestcell}\bfseries 86.46 & \cellcolor{bestcell}\bfseries 81.87 & \cellcolor{bestcell}\bfseries 85.19 & \cellcolor{bestcell}\bfseries 83.88 & \cellcolor{bestcell}\bfseries 85.20 & \cellcolor{bestcell}\bfseries 84.66 & \cellcolor{bestcell}\bfseries 84.50 \\
Gemini-3.1-Pro & 64.18 & 40.84 & 21.13 & 49.45 & 60.32 & 32.33 & 13.30 & 43.00 & 67.89 & 44.30 & 18.69 & 52.14 & 60.91 & 31.97 & 19.10 & 44.02 \\
GPT-5.4 & 60.82 & 64.23 & 60.92 & 62.20 & 69.67 & 65.64 & 59.57 & 66.84 & 69.59 & 67.27 & 58.66 & 67.32 & 70.15 & 66.61 & 56.99 & 67.13 \\
\bottomrule
\end{tabular*}
\caption{Reasoning accuracy (\%) across input settings. Green marks the best result per column; bold marks the second-best. Instruct suffixes are omitted; full names appear in Table~\ref{tab:model_comparison}.}
\label{tab:model_math_comparison}
}
\end{table*}
\begin{figure}[t]
    \centering
    \begin{subfigure}[t]{0.48\linewidth}
        \centering
        \includegraphics[width=\linewidth]{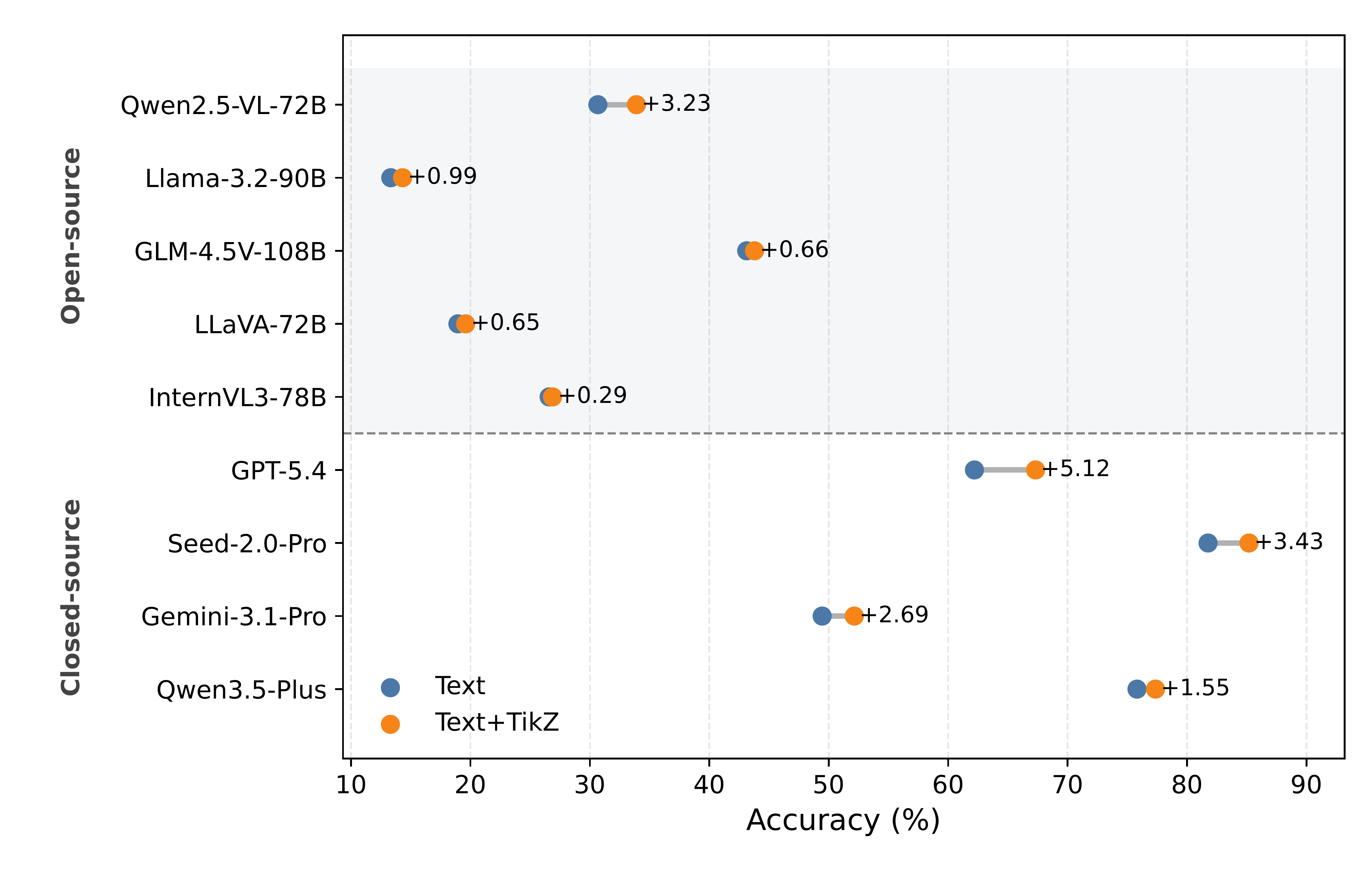}
        \caption{Text vs. Text+\texttt{TikZ}.}
        \label{fig:dumbbell}
    \end{subfigure}
    \hfill
    \begin{subfigure}[t]{0.48\linewidth}
        \centering
        \includegraphics[width=\linewidth]{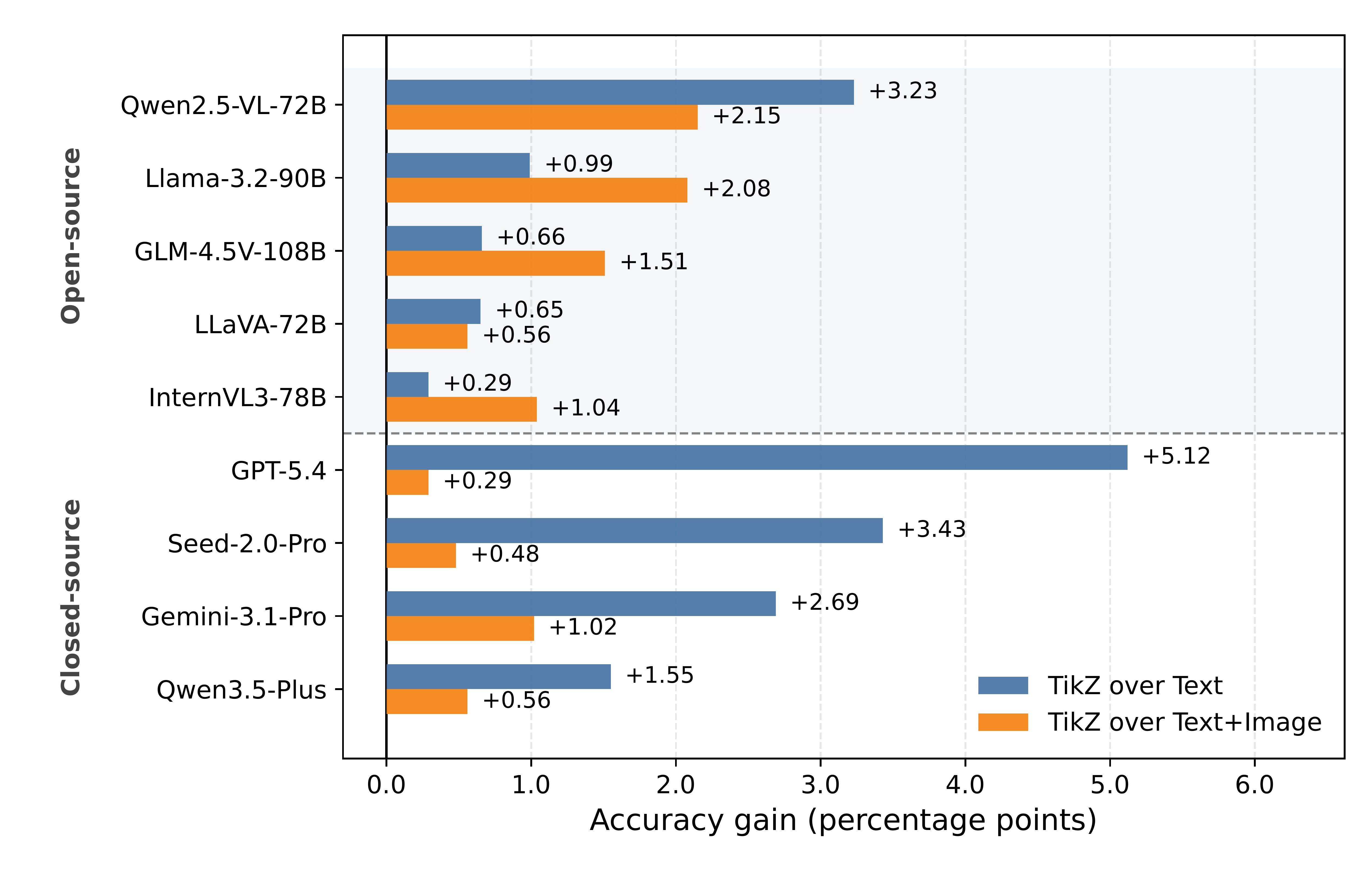}
        \caption{Model-wise \texttt{TikZ} gains.}
        \label{fig:tikz_gain}
    \end{subfigure}
    \par\smallskip
    \begin{subfigure}[t]{0.48\linewidth}
        \centering
        \includegraphics[width=\linewidth]{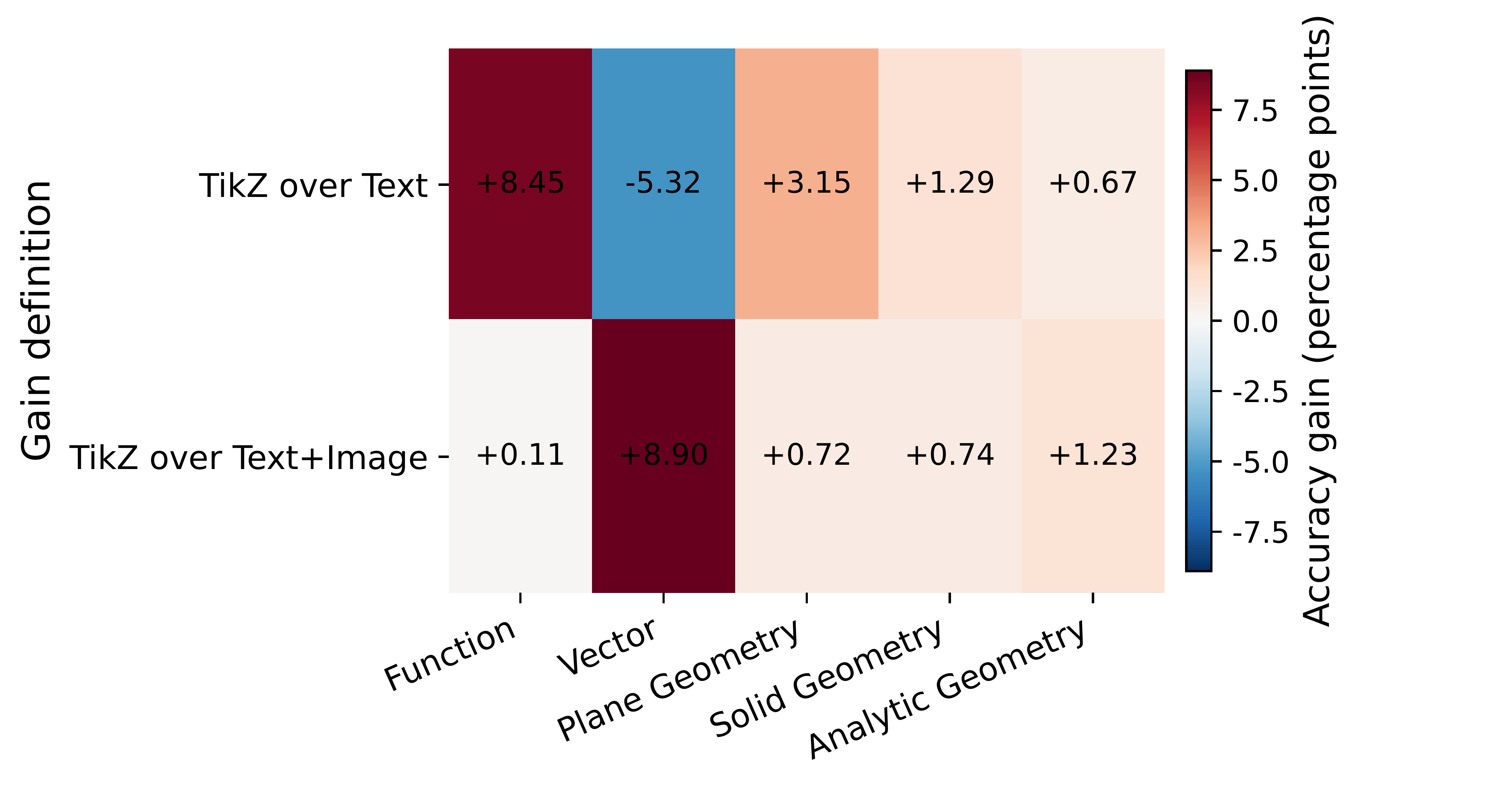}
        \caption{By knowledge point.}
        \label{fig:know_heat}
    \end{subfigure}
    \hfill
    \begin{subfigure}[t]{0.48\linewidth}
        \centering
        \includegraphics[width=\linewidth]{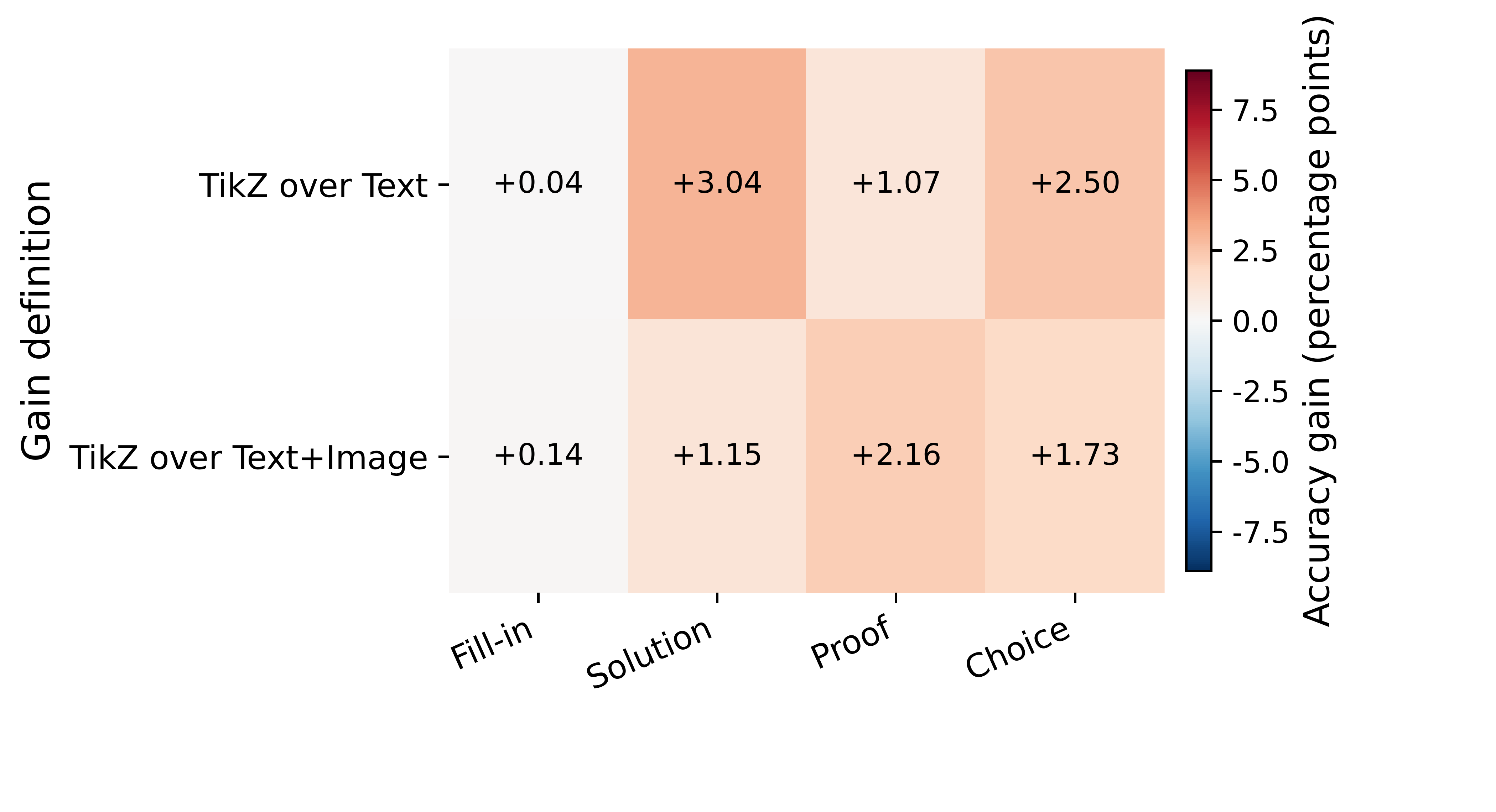}
        \caption{By question type.}
        \label{fig:question_heat}
    \end{subfigure}
    \caption{Oracle effect of human-curated reference \texttt{TikZ}: (a) Text-only versus Text+\texttt{TikZ}; (b) point differences over Text-only and Text+Figure; and (c--d) differences by knowledge point and question type.}
    \label{fig:modelwise_tikz_effect}
    \label{fig:tikz_gain_breakdown}
\end{figure}

Following the LLM-as-Judge paradigm~\cite{li2024llmsasjudgescomprehensivesurveyllmbased}, we use a fixed grading prompt with item-level rules and JSON outputs. The maximum score equals the number of required answers, with one point per correct answer. Human-audit scores are normalized by this maximum. A human audit of 300 responses (78 choice, 84 fill-in, 53 proof, and 85 solution items) yields 91.7\% exact and 99.0\% within-one-point agreement, Pearson $r=0.629$, Spearman $\rho=0.607$, and normalized-score MAE $=0.062$; details are in the appendices.

Table~\ref{tab:model_math_comparison} reports reasoning accuracy under four input settings: Text-only, Text+Figure, Text+\texttt{TikZ}, and Text+Figure+\texttt{TikZ}. 
Overall, closed-source models have higher point estimates across most settings. Seed-2.0-Pro performs best overall, while GPT-5.4 benefits from reference \texttt{TikZ}. Among open-source models, GLM-4.5V-108B performs best on average, whereas Llama-3.2-90B and LLaVA-OneVision-72B remain weaker. These comparisons indicate limited open-source performance on Chinese geometry-intensive reasoning under our setting, rather than a language-controlled estimate of geometry ability.

Seed-2.0-Pro's 81.76\% Text-only score does not contradict figure-grounding: some items contain redundant text, while priors or prior exposure may support correct answers. Aggregate accuracy cannot establish image necessity.

Gemini-3.1-Pro performs best on image-to-\texttt{TikZ} but decreases from 49.45 with Text-only to 43.00 with Text+Figure in reasoning. Image-to-\texttt{TikZ} emphasizes reconstruction, code generation, and format adherence, whereas reasoning also depends on Chinese comprehension, answer formatting, and grader parsing. The decrease may reflect multimodal-input or evaluation-format sensitivity; we do not attribute it solely to geometry reasoning.

Figure~\ref{fig:modelwise_tikz_effect} compares model-wise point estimates. In Text+\texttt{TikZ}, each model receives the reference program associated with the problem; no program is generated from the test image at inference time. Text+\texttt{TikZ} is higher than Text-only for every model, but several open-source differences are below one percentage point and are not statistically conclusive. Adding reference code to Text+Figure also produces positive differences. Because the oracle code can expose exact construction parameters, these are upper-bound diagnostics, not deployable-converter gains.

\begin{table}[t]
    \centering
    \setlength{\tabcolsep}{4.5pt}
    \begin{tabular}{lrrr}
        \toprule
        Benchmark & Text+Fig. & Text+\texttt{TikZ} & Gain \\
        \midrule
        MathVerse   & 57 & \textbf{70} & +13 \\
        MATH-Vision & 69 & \textbf{78} & +9 \\
        VisioMath   & 81 & \textbf{84} & +3 \\
        \bottomrule
    \end{tabular}
    \caption{Exploratory Seed-2.0-Pro accuracy (\%) on 100 samples per benchmark. \texttt{TikZ} is human-authored and verified; Gain is Text+\texttt{TikZ} minus Text+Figure.}
    \label{tab:cross_dataset_tikz}

\end{table}

\noindent \textbf{Exploratory cross-dataset check.}
On 100 samples from each of MathVerse, MATH-Vision, and VisioMath, Table~\ref{tab:cross_dataset_tikz} reports Seed-2.0-Pro increases from 57\% to 70\%, 69\% to 78\%, and 81\% to 84\% with human-verified Text+\texttt{TikZ}. These exploratory results do not evaluate automatic conversion or establish robust generalization, and the VisioMath gain is modest.

Figure~\ref{fig:tikz_gain_breakdown} further breaks down the gains by knowledge point and question type. 
Figure~\ref{fig:know_heat} reports positive point differences across most knowledge areas, especially geometry-related categories where relations among points, lines, curves, and spatial configurations are central. 
Figure~\ref{fig:question_heat} shows larger differences for solution- and proof-style problems than short-answer cases, suggesting potential value for multi-step geometric reasoning.

However, the benefits of additional modalities are not always additive. 
Although oracle Text+\texttt{TikZ} has higher point estimates than Text-only, Text+Figure+\texttt{TikZ} does not always achieve the best absolute accuracy for every model. 
This indicates that current MLLMs still struggle to fuse redundant visual and symbolic inputs. 
Thus, TriGeoBench provides a diagnostic of structured representations and exposes the unresolved challenge of reliable text--figure--structure fusion.
\section{Conclusion}
\textbf{TriGeoBench} represents geometry problems as aligned question--answer, image, and \texttt{TikZ} triplets. Its unified probes expose limitations in structure understanding and image--text alignment, while oracle results with human-curated reference \texttt{TikZ} suggest that accurate executable representations can provide complementary geometric cues.

\section*{Limitations} TriGeoBench focuses on Chinese secondary-school problems, so scores conflate Chinese comprehension, mathematical reasoning, vision, and structured-input use and may not transfer to other languages or curricula. Public sources and unknown proprietary pretraining corpora prevent a complete contamination guarantee; deduplication only reduces obvious overlap. Reference \texttt{TikZ} may expose exact construction parameters, and equivalent programs complicate automatic evaluation. Future work evaluates automatic conversion pipelines and task-specific systems, expands the dataset's scope, and strengthens structure-aware evaluation.

\bibliography{references}

@inproceedings{NEURIPS_DATASETS_AND_BENCHMARKS2021_be83ab3e,
 author = {Hendrycks, Dan and Burns, Collin and Kadavath, Saurav and Arora, Akul and Basart, Steven and Tang, Eric and Song, Dawn and Steinhardt, Jacob},
 booktitle = {Proceedings of the Neural Information Processing Systems Track on Datasets and Benchmarks},
 editor = {J. Vanschoren and S. Yeung},
 pages = {},
 title = {Measuring Mathematical Problem Solving With the MATH Dataset},
 url = {https://datasets-benchmarks-proceedings.neurips.cc/paper_files/paper/2021/file/be83ab3ecd0db773eb2dc1b0a17836a1-Paper-round2.pdf},
 volume = {1},
 year = {2021}
}

@inproceedings{lu2024mathvista,
  title={MathVista: Evaluating Mathematical Reasoning of Foundation Models in Visual Contexts},
  author={Lu, Pan and Bansal, Hritik and Xia, Tony and Liu, Jiacheng and Li, Chunyuan and Hajishirzi, Hannaneh and Cheng, Hao and Chang, Kai-Wei and Galley, Michel and Gao, Jianfeng},
  booktitle={International Conference on Learning Representations (ICLR)},
  year={2024}
}

@inproceedings{
wang2024measuring,
title={Measuring Multimodal Mathematical Reasoning with MATH-Vision Dataset},
author={Ke Wang and Junting Pan and Weikang Shi and Zimu Lu and Houxing Ren and Aojun Zhou and Mingjie Zhan and Hongsheng Li},
booktitle={The Thirty-eight Conference on Neural Information Processing Systems Datasets and Benchmarks Track},
year={2024},
url={https://openreview.net/forum?id=QWTCcxMpPA}
}

@inproceedings{li2026visiomath,
  title     = {VisioMath: Benchmarking Figure-based Mathematical Reasoning in LMMs},
  author    = {Li, Can and Liu, Ying and Zhang, Ting and Wang, Mei and Huang, Hua},
  booktitle = {International Conference on Learning Representations},
  year      = {2026}
}

@inproceedings{zhang-etal-2024-geoeval,
    title = "{G}eo{E}val: Benchmark for Evaluating {LLM}s and Multi-Modal Models on Geometry Problem-Solving",
    author = "Zhang, Jiaxin  and
      Li, Zhong-Zhi  and
      Zhang, Ming-Liang  and
      Yin, Fei  and
      Liu, Cheng-Lin  and
      Moshfeghi, Yashar",
    editor = "Ku, Lun-Wei  and
      Martins, Andre  and
      Srikumar, Vivek",
    booktitle = "Findings of the Association for Computational Linguistics: ACL 2024",
    month = aug,
    year = "2024",
    address = "Bangkok, Thailand",
    publisher = "Association for Computational Linguistics",
    url = "https://aclanthology.org/2024.findings-acl.73/",
    doi = "10.18653/v1/2024.findings-acl.73",
    pages = "1258--1276"
}

@inproceedings{lu2021inter,
 title = {Inter-GPS: Interpretable Geometry Problem Solving with Formal Language and Symbolic Reasoning},
 author = {Lu, Pan and Gong, Ran and Jiang, Shibiao and Qiu, Liang and Huang, Siyuan and Liang, Xiaodan and Zhu, Song-Chun},
 booktitle = {The Joint Conference of the 59th Annual Meeting of the Association for Computational Linguistics and the 11th International Joint Conference on Natural Language Processing (ACL-IJCNLP 2021)},
 year = {2021}
}

@inproceedings{chen-etal-2021-geoqa,
    title = "{G}eo{QA}: A Geometric Question Answering Benchmark Towards Multimodal Numerical Reasoning",
    author = "Chen, Jiaqi  and
      Tang, Jianheng  and
      Qin, Jinghui  and
      Liang, Xiaodan  and
      Liu, Lingbo  and
      Xing, Eric  and
      Lin, Liang",
    editor = "Zong, Chengqing  and
      Xia, Fei  and
      Li, Wenjie  and
      Navigli, Roberto",
    booktitle = "Findings of the Association for Computational Linguistics: ACL-IJCNLP 2021",
    month = aug,
    year = "2021",
    address = "Online",
    publisher = "Association for Computational Linguistics",
    url = "https://aclanthology.org/2021.findings-acl.46/",
    doi = "10.18653/v1/2021.findings-acl.46",
    pages = "513--523"
}

@inproceedings{DBLP:conf/coling/CaoX22,
  author       = {Jie Cao and
                  Jing Xiao},
  editor       = {Nicoletta Calzolari and
                  Chu{-}Ren Huang and
                  Hansaem Kim and
                  James Pustejovsky and
                  Leo Wanner and
                  Key{-}Sun Choi and
                  Pum{-}Mo Ryu and
                  Hsin{-}Hsi Chen and
                  Lucia Donatelli and
                  Heng Ji and
                  Sadao Kurohashi and
                  Patrizia Paggio and
                  Nianwen Xue and
                  Seokhwan Kim and
                  Younggyun Hahm and
                  Zhong He and
                  Tony Kyungil Lee and
                  Enrico Santus and
                  Francis Bond and
                  Seung{-}Hoon Na},
  title        = {An Augmented Benchmark Dataset for Geometric Question Answering through
                  Dual Parallel Text Encoding},
  booktitle    = {Proceedings of the 29th International Conference on Computational
                  Linguistics, {COLING} 2022, Gyeongju, Republic of Korea, October 12-17,
                  2022},
  pages        = {1511--1520},
  publisher    = {International Committee on Computational Linguistics},
  year         = {2022},
  url          = {https://aclanthology.org/2022.coling-1.130},
  bibsource    = {dblp computer science bibliography, https://dblp.org}
}

@inproceedings{chen-etal-2022-unigeo,
    title = "{U}ni{G}eo: Unifying Geometry Logical Reasoning via Reformulating Mathematical Expression",
    author = "Chen, Jiaqi  and
      Li, Tong  and
      Qin, Jinghui  and
      Lu, Pan  and
      Lin, Liang  and
      Chen, Chongyu  and
      Liang, Xiaodan",
    editor = "Goldberg, Yoav  and
      Kozareva, Zornitsa  and
      Zhang, Yue",
    booktitle = "Proceedings of the 2022 Conference on Empirical Methods in Natural Language Processing",
    month = dec,
    year = "2022",
    address = "Abu Dhabi, United Arab Emirates",
    publisher = "Association for Computational Linguistics",
    url = "https://aclanthology.org/2022.emnlp-main.218/",
    doi = "10.18653/v1/2022.emnlp-main.218",
    pages = "3313--3323"
}

@inproceedings{ijcai2023p0376,
  title     = {A Multi-Modal Neural Geometric Solver with Textual Clauses Parsed from Diagram},
  author    = {Zhang, Ming-Liang and yin, Fei and Liu, Cheng-Lin},
  booktitle = {Proceedings of the Thirty-Second International Joint Conference on
               Artificial Intelligence, {IJCAI-23}},
  publisher = {International Joint Conferences on Artificial Intelligence Organization},
  editor    = {Edith Elkind},
  pages     = {3374--3382},
  year      = {2023},
  month     = {8},
  note      = {Main Track},
  doi       = {10.24963/ijcai.2023/376},
  url       = {https://doi.org/10.24963/ijcai.2023/376},
}

@inproceedings{NEURIPS2020_bcf9d6bd,
 author = {Carlier, Alexandre and Danelljan, Martin and Alahi, Alexandre and Timofte, Radu},
 booktitle = {Advances in Neural Information Processing Systems},
 editor = {H. Larochelle and M. Ranzato and R. Hadsell and M.F. Balcan and H. Lin},
 pages = {16351--16361},
 publisher = {Curran Associates, Inc.},
 title = {DeepSVG: A Hierarchical Generative Network for Vector Graphics Animation},
 url = {https://proceedings.neurips.cc/paper_files/paper/2020/file/bcf9d6bd14a2095866ce8c950b702341-Paper.pdf},
 volume = {33},
 year = {2020}
}

@inproceedings{zou-etal-2024-vgbench,
    title = "{VGB}ench: Evaluating Large Language Models on Vector Graphics Understanding and Generation",
    author = "Zou, Bocheng  and
      Cai, Mu  and
      Zhang, Jianrui  and
      Lee, Yong Jae",
    editor = "Al-Onaizan, Yaser  and
      Bansal, Mohit  and
      Chen, Yun-Nung",
    booktitle = "Proceedings of the 2024 Conference on Empirical Methods in Natural Language Processing",
    month = nov,
    year = "2024",
    address = "Miami, Florida, USA",
    publisher = "Association for Computational Linguistics",
    url = "https://aclanthology.org/2024.emnlp-main.213/",
    doi = "10.18653/v1/2024.emnlp-main.213",
    pages = "3647--3659"
}

@inproceedings{belouadi2024detikzify,
    title={{DeTikZify}: Synthesizing Graphics Programs for Scientific Figures and Sketches with {TikZ}},
    author={Jonas Belouadi and Simone Paolo Ponzetto and Steffen Eger},
    booktitle={The Thirty-eighth Annual Conference on Neural Information Processing Systems},
    year={2024},
    url={https://openreview.net/forum?id=bcVLFQCOjc}
}

@article{luo2025geogrambench,
  title={Geogrambench: Benchmarking the geometric program reasoning in modern llms},
  author={Luo, Shixian and Zhu, Zezhou and Yuan, Yu and Yang, Yuncheng and Shan, Lianlei and Wu, Yong},
  journal={arXiv preprint arXiv:2505.17653},
  year={2025}
}

@article{chen2025symbolic,
  title={Symbolic Graphics Programming with Large Language Models}, 
  author={Yamei Chen and Haoquan Zhang and Yangyi Huang and Zeju Qiu and Kaipeng Zhang and Yandong Wen and Weiyang Liu},
  journal={arXiv preprint arXiv:2509.05208},
  year={2025}
}

@article{yang2025omnisvg,
  title={OmniSVG: A Unified Scalable Vector Graphics Generation Model},
  author={Yang, Yiying and Cheng, Wei and Chen, Sijin and Zeng, Xianfang and Zhang, Jiaxu and Wang, Liao and Yu, Gang and Ma, Xingjun and Jiang, Yu-Gang},
  journal={arXiv preprint arXiv:2504.06263},
  year={2025}
}

@article{wang2025internsvg,
    title={InternSVG: Towards Unified SVG Tasks with Multimodal Large Language Models},
    author={Wang, Haomin and Yin, Jinhui and Wei, Qi and Zeng, Wenguang and Gu, Lixin and Ye, Shenglong and Gao, Zhangwei
    and Wang, Yaohui and Zhang, Yanting and Li, Yuanqi and others},
    journal={arXiv preprint arXiv:2510.11341},
    year={2025}
}

@inproceedings{10.1145/3711896.3737436,
author = {Gervais, Philippe and Fadeeva, Anastasiia and Maksai, Andrii},
title = {MathWriting: A Dataset For Handwritten Mathematical Expression Recognition},
year = {2025},
isbn = {9798400714542},
publisher = {Association for Computing Machinery},
address = {New York, NY, USA},
url = {https://doi.org/10.1145/3711896.3737436},
doi = {10.1145/3711896.3737436},
booktitle = {Proceedings of the 31st ACM SIGKDD Conference on Knowledge Discovery and Data Mining V.2},
pages = {5459–5469},
numpages = {11},
location = {Toronto ON, Canada},
series = {KDD '25}
}

@inproceedings{10.1007/978-3-031-73242-3_10,
author = {Zhang, Renrui and Jiang, Dongzhi and Zhang, Yichi and Lin, Haokun and Guo, Ziyu and Qiu, Pengshuo and Zhou, Aojun and Lu, Pan and Chang, Kai-Wei and Qiao, Yu and Gao, Peng and Li, Hongsheng},
title = {MATHVERSE: Does Your Multi-modal LLM Truly See the\&nbsp;Diagrams in\&nbsp;Visual Math Problems?},
year = {2024},
isbn = {978-3-031-73241-6},
publisher = {Springer-Verlag},
address = {Berlin, Heidelberg},
url = {https://doi.org/10.1007/978-3-031-73242-3_10},
doi = {10.1007/978-3-031-73242-3_10},
booktitle = {Computer Vision – ECCV 2024: 18th European Conference, Milan, Italy, September 29–October 4, 2024, Proceedings, Part VIII},
pages = {169–186},
numpages = {18},
location = {Milan, Italy}
}

@article{liu2023visual,
  title={Visual instruction tuning},
  author={Liu, Haotian and Li, Chunyuan and Wu, Qingyang and Lee, Yong Jae},
  journal={Advances in neural information processing systems},
  volume={36},
  pages={34892--34916},
  year={2023}
}

@article{zhu2023minigpt,
  title={Minigpt-4: Enhancing vision-language understanding with advanced large language models},
  author={Zhu, Deyao and Chen, Jun and Shen, Xiaoqian and Li, Xiang and Elhoseiny, Mohamed},
  journal={arXiv preprint arXiv:2304.10592},
  year={2023}
}

@article{hurst2024gpt,
  title={Gpt-4o system card},
  author={Hurst, Aaron and Lerer, Adam and Goucher, Adam P and Perelman, Adam and Ramesh, Aditya and Clark, Aidan and Ostrow, AJ and Welihinda, Akila and Hayes, Alan and Radford, Alec and others},
  journal={arXiv preprint arXiv:2410.21276},
  year={2024}
}

@article{team2023gemini,
  title={Gemini: a family of highly capable multimodal models},
  author={Team, Gemini and Anil, Rohan and Borgeaud, Sebastian and Alayrac, Jean-Baptiste and Yu, Jiahui and Soricut, Radu and Schalkwyk, Johan and Dai, Andrew M and Hauth, Anja and Millican, Katie and others},
  journal={arXiv preprint arXiv:2312.11805},
  year={2023}
}

@inproceedings{li2023blip,
  title={Blip-2: Bootstrapping language-image pre-training with frozen image encoders and large language models},
  author={Li, Junnan and Li, Dongxu and Savarese, Silvio and Hoi, Steven},
  booktitle={International conference on machine learning},
  pages={19730--19742},
  year={2023},
  organization={PMLR}
}

@inproceedings{liu2024mmbench,
  title={Mmbench: Is your multi-modal model an all-around player?},
  author={Liu, Yuan and Duan, Haodong and Zhang, Yuanhan and Li, Bo and Zhang, Songyang and Zhao, Wangbo and Yuan, Yike and Wang, Jiaqi and He, Conghui and Liu, Ziwei and others},
  booktitle={European conference on computer vision},
  pages={216--233},
  year={2024},
  organization={Springer}
}

@inproceedings{hudson2019gqa,
  title={Gqa: A new dataset for real-world visual reasoning and compositional question answering},
  author={Hudson, Drew A and Manning, Christopher D},
  booktitle={Proceedings of the IEEE/CVF conference on computer vision and pattern recognition},
  pages={6700--6709},
  year={2019}
}

@inproceedings{masry2022chartqa,
  title={Chartqa: A benchmark for question answering about charts with visual and logical reasoning},
  author={Masry, Ahmed and Do, Xuan Long and Tan, Jia Qing and Joty, Shafiq and Hoque, Enamul},
  booktitle={Findings of the association for computational linguistics: ACL 2022},
  pages={2263--2279},
  year={2022}
}

@article{hendrycks2021measuring,
  title={Measuring mathematical problem solving with the math dataset},
  author={Hendrycks, Dan and Burns, Collin and Kadavath, Saurav and Arora, Akul and Basart, Steven and Tang, Eric and Song, Dawn and Steinhardt, Jacob},
  journal={arXiv preprint arXiv:2103.03874},
  year={2021}
}

@article{yang2024askchart,
  title={Askchart: Universal chart understanding through textual enhancement},
  author={Yang, Xudong and Wu, Yifan and Zhu, Yizhang and Tang, Nan and Luo, Yuyu},
  journal={arXiv preprint arXiv:2412.19146},
  year={2024}
}

@inproceedings{mathew2021docvqa,
  title={Docvqa: A dataset for vqa on document images},
  author={Mathew, Minesh and Karatzas, Dimosthenis and Jawahar, CV},
  booktitle={Proceedings of the IEEE/CVF winter conference on applications of computer vision},
  pages={2200--2209},
  year={2021}
}

@inproceedings{tang2025mtvqa,
  title={Mtvqa: Benchmarking multilingual text-centric visual question answering},
  author={Tang, Jingqun and Liu, Qi and Ye, Yongjie and Lu, Jinghui and Wei, Shu and Wang, An-Lan and Lin, Chunhui and Feng, Hao and Zhao, Zhen and Wang, Yanjie and others},
  booktitle={Findings of the Association for Computational Linguistics: ACL 2025},
  pages={7748--7763},
  year={2025}
}

@article{lu2022learn,
  title={Learn to explain: Multimodal reasoning via thought chains for science question answering},
  author={Lu, Pan and Mishra, Swaroop and Xia, Tanglin and Qiu, Liang and Chang, Kai-Wei and Zhu, Song-Chun and Tafjord, Oyvind and Clark, Peter and Kalyan, Ashwin},
  journal={Advances in neural information processing systems},
  volume={35},
  pages={2507--2521},
  year={2022}
}

@inproceedings{yue2025mmmu,
  title={Mmmu-pro: A more robust multi-discipline multimodal understanding benchmark},
  author={Yue, Xiang and Zheng, Tianyu and Ni, Yuansheng and Wang, Yubo and Zhang, Kai and Tong, Shengbang and Sun, Yuxuan and Yu, Botao and Zhang, Ge and Sun, Huan and others},
  booktitle={Proceedings of the 63rd Annual Meeting of the Association for Computational Linguistics (Volume 1: Long Papers)},
  pages={15134--15186},
  year={2025}
}

@article{li2024survey,
  title={A survey on benchmarks of multimodal large language models},
  author={Li, Jian and Lu, Weiheng and Fei, Hao and Luo, Meng and Dai, Ming and Xia, Min and Jin, Yizhang and Gan, Zhenye and Qi, Ding and Fu, Chaoyou and others},
  journal={arXiv preprint arXiv:2408.08632},
  year={2024}
}

@article{caffagni2024revolution,
  title={The revolution of multimodal large language models: A survey},
  author={Caffagni, Davide and Cocchi, Federico and Barsellotti, Luca and Moratelli, Nicholas and Sarto, Sara and Baraldi, Lorenzo and Cornia, Marcella and Cucchiara, Rita},
  journal={Findings of the association for computational linguistics: ACL 2024},
  pages={13590--13618},
  year={2024}
}

@article{wang2025solidgeo,
  title={Solidgeo: Measuring multimodal spatial math reasoning in solid geometry},
  author={Wang, Peijie and Yang, Chao and Li, Zhong-Zhi and Yin, Fei and Ran, Dekang and Tian, Mi and Ji, Zhilong and Bai, Jinfeng and Liu, Cheng-Lin},
  journal={arXiv preprint arXiv:2505.21177},
  year={2025}
}

@article{cobbe2021training,
  title={Training verifiers to solve math word problems},
  author={Cobbe, Karl and Kosaraju, Vineet and Bavarian, Mohammad and Chen, Mark and Jun, Heewoo and Kaiser, Lukasz and Plappert, Matthias and Tworek, Jerry and Hilton, Jacob and Nakano, Reiichiro and others},
  journal={arXiv preprint arXiv:2110.14168},
  year={2021}
}

@article{gao2024omni,
  title={Omni-math: A universal olympiad level mathematic benchmark for large language models},
  author={Gao, Bofei and Song, Feifan and Yang, Zhe and Cai, Zefan and Miao, Yibo and Dong, Qingxiu and Li, Lei and Ma, Chenghao and Chen, Liang and Xu, Runxin and others},
  journal={arXiv preprint arXiv:2410.07985},
  year={2024}
}

@misc{chatgpt54,
    author={OpenAI},
    title={GPT-5.4},
    year={2026},
    howpublished={\url{https://openai.com/index/introducing-gpt-5-4/}},
}

@article{bai2025qwen3,
  title={Qwen3-vl technical report},
  author={Bai, Shuai and Cai, Yuxuan and others},
  journal={arXiv preprint arXiv:2511.21631},
  year={2025}
}

@article{bai2025qwen25,
  title={Qwen2.5-vl technical report},
  author={Bai Shuai, Chen Keqin, Liu Xuejing and others},
  journal={arXiv preprint arXiv:2502.13923},
  year={2025}
}

@article{hong2025glm,
  title={Glm-4.5 v and glm-4.1 v-thinking: Towards versatile multimodal reasoning with scalable reinforcement learning},
  author={Hong, Wenyi and Yu, Wenmeng and Gu, Xiaotao and others},
  journal={arXiv preprint arXiv:2507.01006},
  year={2025}
}

@misc{meta2024llama32hf,
  title        = {Llama-3.2-11B-Vision},
  author       = {{Meta}},
  year         = {2024},
  howpublished = {\url{https://huggingface.co/meta-llama/Llama-3.2-11B-Vision}},
}

@article{gemma2025,
  title={Gemma 3 technical report},
  author={Aishwarya Kamath, Johan Ferret and others},
  journal={arXiv preprint arXiv:2503.19786},
  year={2025}
}

@article{li2024llava,
  title={Llava-onevision: Easy visual task transfer},
  author={Li, Bo and Zhang, Yuanhan and Guo, Dong and Zhang, Renrui and Li, Feng and Zhang, Hao and Zhang, Kaichen and Zhang, Peiyuan and Li, Yanwei and Liu, Ziwei and others},
  journal={arXiv preprint arXiv:2408.03326},
  year={2024}
}

@article{zhu2025internvl3,
  title={Internvl3: Exploring advanced training and test-time recipes for open-source multimodal models},
  author={Zhu, Jinguo and Wang, Weiyun and Chen, Zhe and Liu, Zhaoyang and Ye, Shenglong and Gu, Lixin and Tian, Hao and Duan, Yuchen and Su, Weijie and Shao, Jie and others},
  journal={arXiv preprint arXiv:2504.10479},
  year={2025}
}

@misc{google2026gemini31pro,
  title        = {Gemini 3.1 Pro},
  author       = {{Google DeepMind}},
  year         = {2026},
  howpublished = {\url{https://deepmind.google/models/model-cards/gemini-3-1-pro/}},
}

@misc{bytedance2025doubao15pro,
  title        = {Doubao-1.5-pro / Seed1.5},
  author       = {{ByteDance Seed Team}},
  year         = {2026},
  howpublished = {\url{https://seed.bytedance.com/en/seed2}},
}

@misc{qwen2026qwen35plus,
  title        = {Qwen3.5-Plus},
  author       = {{Qwen Team}},
  year         = {2026},
  howpublished = {\url{https://www.alibabacloud.com/help/en/model-studio/models}},
}

@misc{li2024llmsasjudgescomprehensivesurveyllmbased,
      title={LLMs-as-Judges: A Comprehensive Survey on LLM-based Evaluation Methods}, 
      author={Haitao Li and Qian Dong and Junjie Chen and Huixue Su and Yujia Zhou and Qingyao Ai and Ziyi Ye and Yiqun Liu},
      year={2024},
      eprint={2412.05579},
      archivePrefix={arXiv},
      primaryClass={cs.CL},
      url={https://arxiv.org/abs/2412.05579}, 
}
\appendix
\section{Data Source, Licensing, and Leakage}
\label{app:data-ethics}

\textbf{Data sources.}
TriGeoBench is constructed from publicly accessible Chinese junior- and senior-secondary mathematics examination materials, including online educational platforms, school repositories, and open exam archives. The raw materials typically contain question statements, answer keys, solution analyses, and associated figures. During preprocessing, we remove irrelevant metadata, source-specific headers and footers, question indices that do not affect problem semantics, and any personal or institution-specific information when present.

\textbf{Licensing and release.}
Since the raw examination materials are collected from heterogeneous public sources, their redistribution conditions may vary. We therefore release the processed dataset in a form that follows the applicable licensing constraints of the original sources. When necessary, we provide derived annotations, normalized \LaTeX{} question--answer pairs, rendered figures, \texttt{TikZ} code, and metadata, while avoiding redistribution of source documents that are not explicitly licensed for reuse. The final release includes a data statement describing source categories, preprocessing steps, and usage restrictions.

\textbf{Deduplication and leakage control.}
To reduce duplicate or near-duplicate samples, we conduct filtering over the collected corpus using textual similarity and manual inspection. Problems based on highly redundant templates are removed during annotation. We further check candidate samples against representative public geometry and multimodal mathematical reasoning benchmarks when applicable. However, because some large multimodal models may have been trained on broad web-scale data, we cannot fully rule out prior exposure to public examination problems. We therefore report the dataset construction procedure transparently and encourage future evaluations to consider potential training-data contamination when interpreting results.

\textbf{Privacy and sensitive information.}
The dataset is built from educational mathematics problems rather than personal records. We remove names, school-specific metadata, document headers, and other irrelevant identifying information when they appear in source documents. The retained content focuses on mathematical problem statements, solutions, figures, and structured annotations.

\section{Annotation Details}
\label{app:annotation-details}

\subsection{Collection and Preprocessing}
\label{app:collection-preprocessing}

We collect publicly available Chinese mathematics examination papers from online educational platforms, school repositories, and open exam archives. The collected materials cover both junior and senior secondary education and typically include question statements, answer keys, and detailed solution analyses. These materials provide both the problem context and the intended reasoning process, which helps verify whether extracted figures contain solution-relevant, nondecorative information.

From the raw materials, we extract candidate problems that contain figures and are relevant to one or more target knowledge areas, including plane geometry, solid geometry, analytic geometry, functions, and vector geometry. Each candidate problem is segmented into structured components, including the question stem, answer, solution, and associated figure(s). Due to heterogeneous layouts and document styles, we perform careful normalization, including cleaning line breaks, removing question indices, deleting irrelevant headers and footers, standardizing mathematical notation, and aligning each question with its corresponding figure region.

After preprocessing, annotators screen the candidate pool by removing questions without valid figures, cases with severely degraded visual quality, and problems based on clearly redundant templates. This filtering step improves the quality and diversity of the dataset while ensuring that the retained instances require meaningful text--figure reasoning rather than relying only on textual shortcuts or superficial visual patterns.

\subsection{Manual Conversion to \LaTeX{} and \texttt{TikZ}}
\label{app:manual-conversion}

To obtain unified and editable representations, human annotators manually transcribe each problem statement into \LaTeX{} and author the corresponding figure in \texttt{TikZ} using the original source as reference. Shared annotation guidelines require annotators to preserve mathematical notation and problem semantics while producing complete, compilable code. The full guidelines are provided in the ``Guidelines for Manual Image-to-TikZ Annotation'' section below.

Manual \texttt{TikZ} authoring is particularly demanding because complex figures require precise topology, geometric incidence, symbolic labels, hidden edges, auxiliary lines, and relative positioning. Annotators therefore compile and compare each representation with the original figure throughout the authoring process, with particular attention to solid-geometry diagrams, function plots with critical points, and figures involving multiple nested geometric relations.

\subsection{Independent Verification and Correction}
\label{app:manual-verification}

For every manually converted instance, an independent reviewer compares the transcribed \LaTeX{} and authored \texttt{TikZ} code with the original source materials. The verification process focuses on four recurring categories of errors:

\begin{enumerate}
    \item \textbf{Mathematical notation errors.} These include malformed fractions, missing operators, incorrect symbols, and wrong subscripts or superscripts.
    \item \textbf{Semantic omissions or distortions.} These include missing conditions, incomplete solution steps, incorrect answer transcription, or changes in the meaning of the original problem.
    \item \textbf{Text--figure inconsistencies.} These include mismatched point labels, missing geometric objects, inconsistent references between the question and the figure, or incorrect association between a figure and its problem statement.
    \item \textbf{\texttt{TikZ} structural errors.} These include incorrect topology, missing auxiliary lines, inaccurate relative positions, misplaced labels, missing hidden edges, and incorrect rendering of geometric or functional relations.
\end{enumerate}

Items that do not pass review are returned to the annotators for correction or complete redrawing when necessary. The final \texttt{TikZ} code must be compilable and preserve the mathematical intent and visual reasoning structure of the original figure.

\begin{figure}[!t]
    \centering
    \begin{minipage}{\linewidth}
        \centering
        \includegraphics[width=\linewidth]{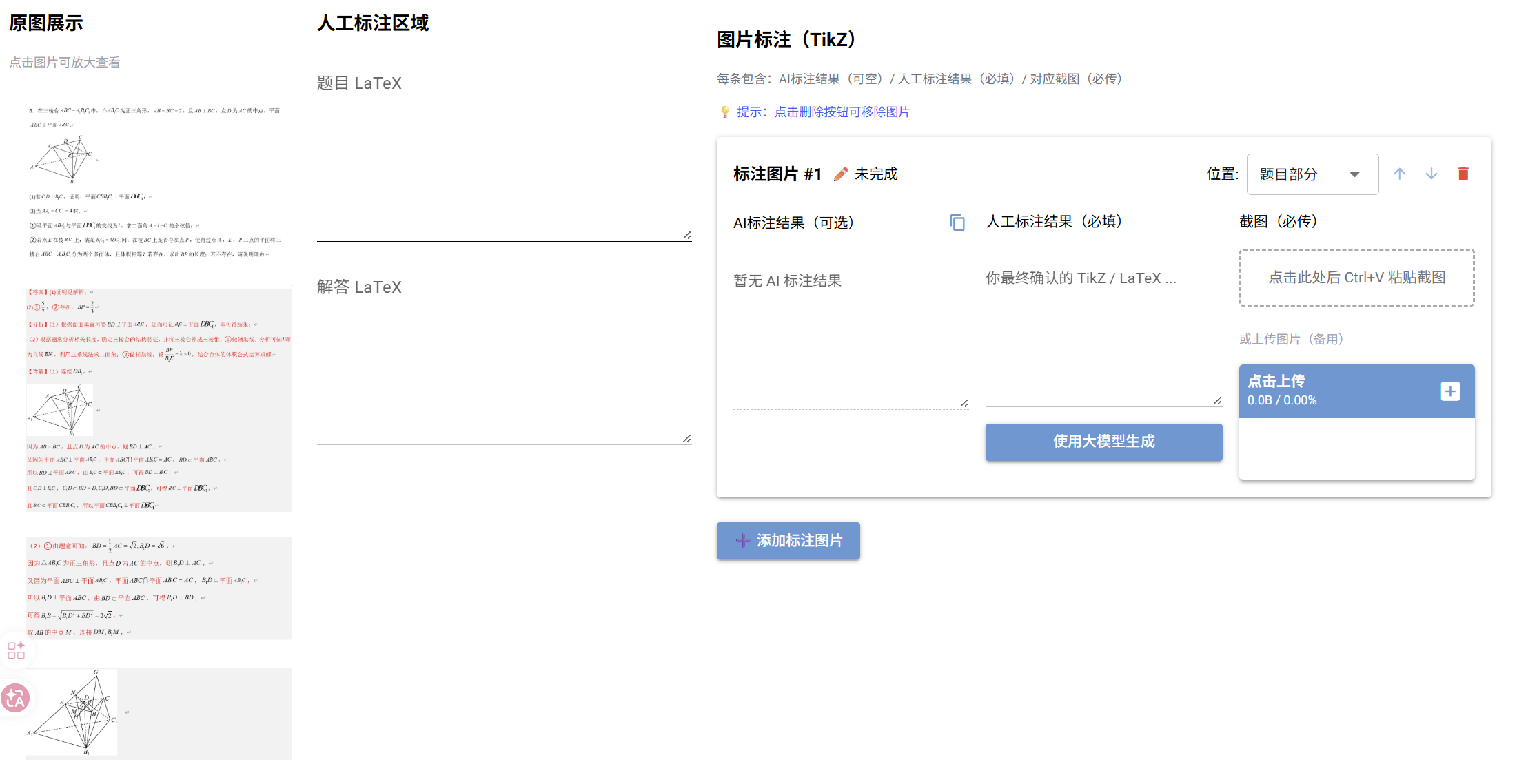}
        \vspace{2pt}
        {\small (a) Annotation interface}
    \end{minipage}
    \par\smallskip
    \begin{minipage}{\linewidth}
        \centering
        \includegraphics[width=\linewidth]{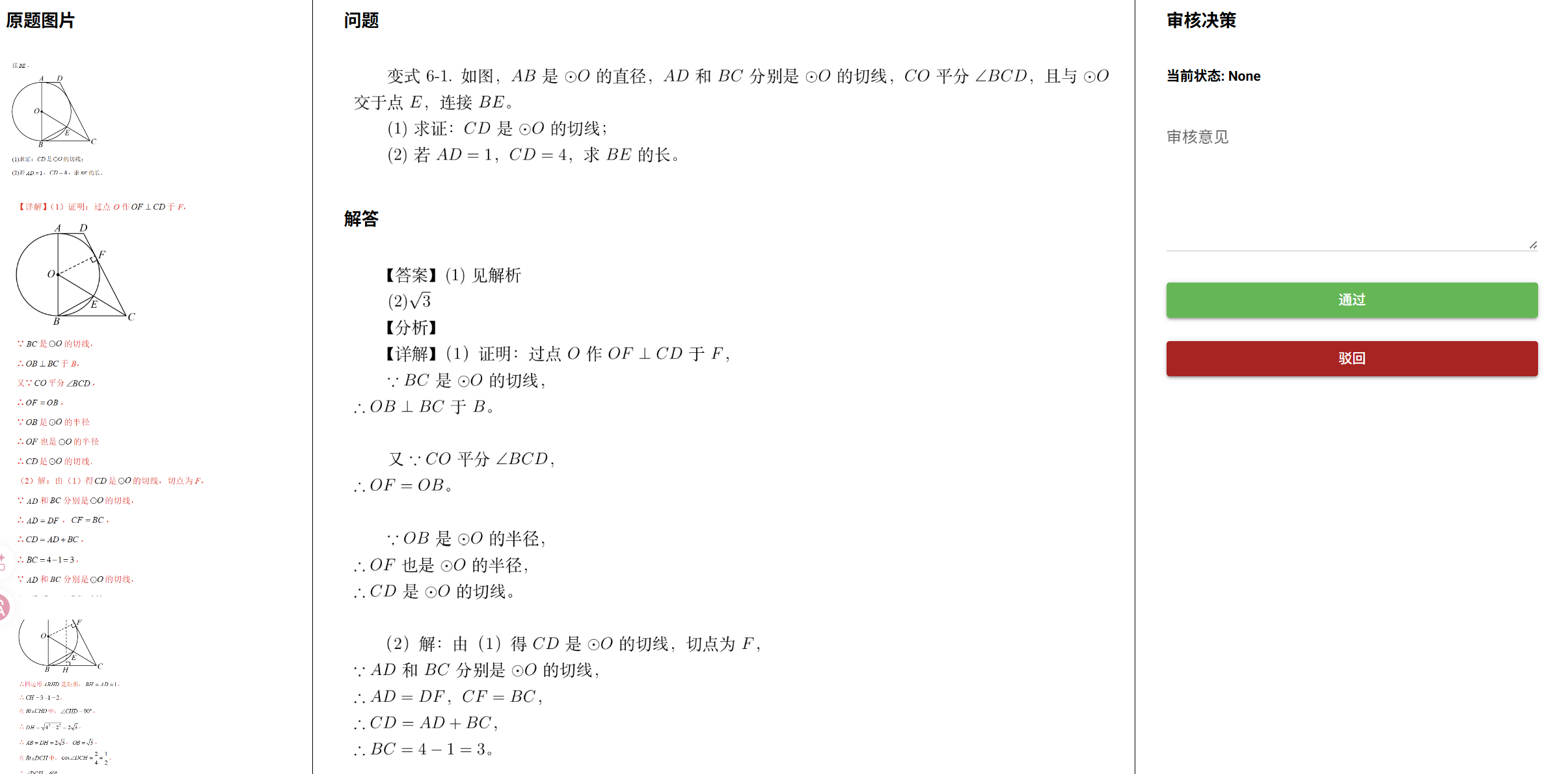}
        \vspace{2pt}
        {\small (b) Review text}
    \end{minipage}
    \par\smallskip
    \begin{minipage}{\linewidth}
        \centering
        \includegraphics[width=\linewidth]{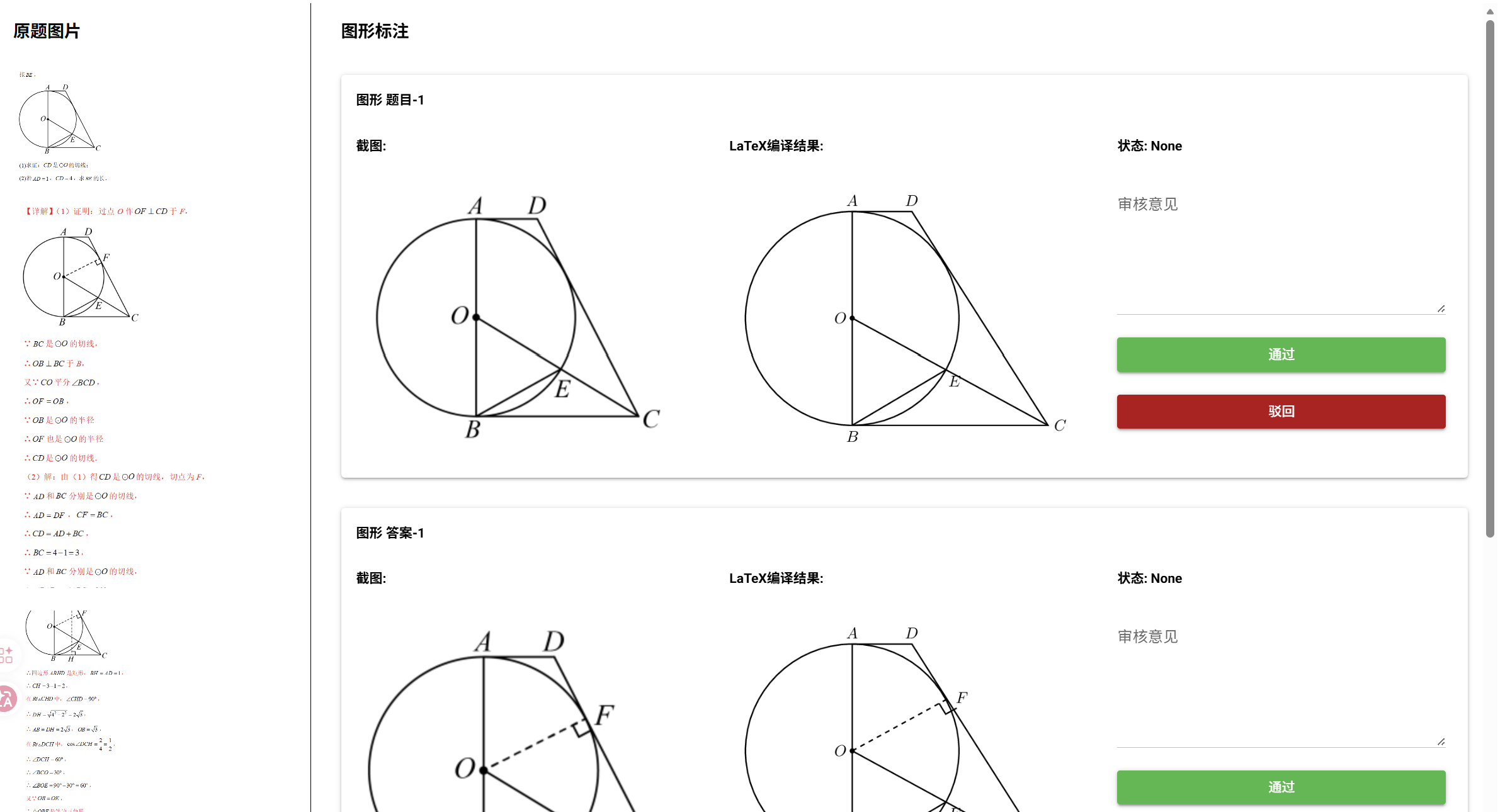}
        \vspace{2pt}
        {\small (c) Review \texttt{TikZ}}
    \end{minipage}
    \caption{Annotation and review interfaces for side-by-side comparison, \LaTeX{}/\texttt{TikZ} editing, and independent verification. English translations and Pinyin of interface labels appear in Appendix~\ref{app:interface-translations}.}
    \label{fig:annotation-system}
\end{figure}

To support large-scale and traceable human annotation, we use an internal annotation and review system, as illustrated in Figure~\ref{fig:annotation-system}. The system presents the original problem image, the manually transcribed \LaTeX{} text, and the authored \texttt{TikZ} code in a unified interface, allowing annotators to compare the source material directly with the structured representation. Annotators can edit the problem statement, solution, answer, and figure code while immediately checking whether the \texttt{TikZ} code compiles and visually matches the original diagram. This interface reduces context switching during annotation and helps annotators focus on mathematical fidelity rather than file-level operations.

The review system provides an additional quality-control layer. Each annotated instance is submitted for inspection, where reviewers check mathematical correctness, text--figure consistency, and rendering quality before accepting the item into the final dataset. Items with unresolved errors are returned for further correction. The system records annotation status and review decisions, enabling us to track the progress of each instance and avoid duplicated or conflicting edits. This system-based workflow makes manual authoring and verification more consistent, auditable, and scalable than independent offline editing.

\FloatBarrier
\section{Evaluation Metrics}
\label{app:evaluation-metrics}

We evaluate the generated \texttt{TikZ}/\LaTeX{} outputs from four complementary perspectives: renderability, visual fidelity, code-level similarity, and human-verified correctness. In addition, for the mathematical reasoning task, we report answer accuracy based on automatic grading by an LLM judge.

\subsection{Render Success Rate}

Render success rate, also referred to as generation accuracy (Gen Acc), measures whether the predicted \texttt{TikZ}/\LaTeX{} code can be successfully compiled into a valid image. Given $N$ test samples and $N_{\mathrm{ok}}$ successfully rendered samples, it is defined as:
\begin{equation}
\mathrm{GenAcc}=\frac{N_{\mathrm{ok}}}{N}.
\end{equation}
This metric evaluates the syntactic correctness and executability of generated code. Since \texttt{TikZ} generation is sensitive to missing environments, invalid commands, and syntax errors, Gen Acc is an important indicator of practical usability.

\subsection{Visual Similarity Metrics}

To evaluate whether the rendered image from generated code matches the ground-truth figure, we use both low-level and high-level visual metrics.

\paragraph{SSIM.}
The Structural Similarity Index Measure (SSIM) measures local structural consistency between the predicted image $x$ and the ground-truth image $y$:
\begin{equation}
\mathrm{SSIM}(x,y)=
\frac{(2\mu_x\mu_y+C_1)(2\sigma_{xy}+C_2)}
{(\mu_x^2+\mu_y^2+C_1)(\sigma_x^2+\sigma_y^2+C_2)}.
\end{equation}
Here, $\mu$, $\sigma^2$, and $\sigma_{xy}$ denote image means, variances, and covariance, respectively. SSIM captures luminance, contrast, and structural similarity, making it suitable for checking whether lines, shapes, and local geometric patterns are preserved.

\paragraph{CLIP and DINO similarity.}
We further compute feature-level cosine similarity using CLIP and DINOv2 visual encoders:
\begin{equation}
\mathrm{Sim}(x,y)=
\frac{f_x^\top f_y}{\|f_x\|_2\|f_y\|_2},
\end{equation}
where $f_x$ and $f_y$ denote visual embeddings of the predicted and reference images. CLIP similarity reflects high-level semantic alignment, while DINO similarity is more sensitive to visual layout and geometric structure, which is particularly useful for mathematical diagrams.

\paragraph{KID.}
Kernel Inception Distance (KID) measures dataset-level distributional similarity between generated and reference images. It is based on the squared Maximum Mean Discrepancy:
\begin{equation}
\begin{aligned}
\mathrm{MMD}^2(X,Y)={}&\frac{1}{m^2}\sum_{i,j}k(x_i,x_j)\\
&+\frac{1}{n^2}\sum_{i,j}k(y_i,y_j)\\
&-\frac{2}{mn}\sum_{i,j}k(x_i,y_j),
\end{aligned}
\end{equation}
where $X$ and $Y$ are feature sets extracted from generated and reference images. Unlike pairwise metrics such as SSIM, CLIP, and DINO, KID evaluates whether the overall generated image distribution matches the reference distribution. Lower KID indicates better distributional consistency.

\subsection{Code-level Similarity Metrics}

We evaluate the similarity between generated and reference \texttt{TikZ} code using lexical, token-level, and structural metrics.

\paragraph{Code BLEU.}
Code BLEU (cBLEU) is computed using sentence-level BLEU:
\begin{equation}
\mathrm{BLEU}
=
BP\cdot
\exp\left(
\sum_{n=1}^{N}w_n\log p_n
\right),
\end{equation}
where $p_n$ is the modified $n$-gram precision, $w_n$ is the weight for each $n$-gram order, and $BP$ is the brevity penalty. cBLEU measures lexical and local syntactic overlap between generated and reference code.

\paragraph{Token F1.}
Token F1 measures token-level agreement after normalizing and tokenizing the generated and reference code:
\begin{equation}
\begin{gathered}
P=\frac{TP}{TP+FP}, \qquad R=\frac{TP}{TP+FN},\\
F1=\frac{2PR}{P+R}.
\end{gathered}
\end{equation}
Compared with BLEU, Token F1 is less sensitive to local ordering differences and is useful when semantically similar \texttt{TikZ} programs are written in different syntactic forms.

\paragraph{Tree Edit Distance.}
Tree Edit Distance (TED) measures structural discrepancy between generated and reference \texttt{TikZ} code. We construct lightweight syntax trees by extracting key graphical primitives and commands, such as \texttt{\textbackslash draw}, \texttt{\textbackslash node}, \texttt{\textbackslash path}, \texttt{circle}, \texttt{rectangle}, \texttt{arc}, and \texttt{plot}. TED is defined as:
\begin{equation}
\mathrm{TED}(T_1,T_2)=
\min\sum_i \mathrm{cost}(op_i),
\end{equation}
where $op_i$ denotes insertion, deletion, or substitution operations. Lower TED indicates stronger structural consistency between the predicted and reference programs.

\subsection{Human Evaluation Accuracy}

Automatic metrics cannot fully determine whether a generated diagram is mathematically faithful, especially when different \texttt{TikZ} programs may produce visually equivalent or semantically acceptable figures. Therefore, we additionally report human evaluation accuracy for the Image-to-\texttt{TikZ} task. Human annotators inspect the rendered prediction and judge whether it is visually and semantically consistent with the ground-truth figure. Given $N$ evaluated samples and $N_{\mathrm{pass}}$ samples judged as acceptable, Human Acc is defined as:
\begin{equation}
\mathrm{HumanAcc}=\frac{N_{\mathrm{pass}}}{N}.
\end{equation}
A sample is considered acceptable only if the rendered figure preserves the key mathematical objects, labels, geometric relations, and visual structure required by the original problem.

\subsection{Mathematical Reasoning Accuracy}

For the mathematical reasoning task, we evaluate model answers using an LLM-based grader with a fixed grading prompt. The grader compares the model response against the reference solution and returns a structured judgment. Accuracy is computed as:
\begin{equation}
\mathrm{Acc}=\frac{N_{\mathrm{correct}}}{N_{\mathrm{total}}},
\end{equation}
where $N_{\mathrm{correct}}$ is the number of correctly answered questions and $N_{\mathrm{total}}$ is the total number of evaluated questions. If a problem contains multiple sub-questions, each sub-question is counted as an independent unit. Thus, a model receives credit for each correctly answered sub-question rather than only for the entire problem.

\subsection{Overall Average Score}

For the Image-to-\texttt{TikZ} task, we also report an aggregated average score by normalizing all metrics into a common higher-is-better scale. For example, cBLEU is divided by 100, KID and TED are transformed by reciprocal normalization, and cosine similarities are mapped to $[0,1]$ when necessary:
\begin{equation}
\begin{gathered}
\widetilde{\mathrm{KID}}=\frac{1}{1+\max(\mathrm{KID},0)},\\
\widetilde{\mathrm{TED}}=\frac{1}{1+\max(\mathrm{TED},0)}.
\end{gathered}
\end{equation}
The final average score is:
\begin{equation}
\mathrm{AVG}=
\frac{1}{|\mathcal{M}|}
\sum_{m\in\mathcal{M}}\widetilde{m},
\end{equation}
where $\mathcal{M}$ denotes the set of normalized metrics. This score provides a compact summary of renderability, visual fidelity, code similarity, and structural consistency.

\subsection{Quality Control}
\label{app:quality-control}

To ensure consistency across modalities, each finalized instance is checked from three perspectives. First, the problem text, answer, and solution must preserve the semantics of the original source. Second, the rendered figure from the \texttt{TikZ} code must match the original figure in terms of key geometric objects, labels, topology, and relative spatial relations. Third, textual references in the problem must be aligned with the corresponding visual entities in the figure. These checks ensure that each data instance forms a coherent \{\LaTeX{} question--answer pair, image, \texttt{TikZ}\} triplet suitable for both image-to-\texttt{TikZ} generation and multimodal mathematical reasoning evaluation.

\section{Guidelines for Manual Image-to-TikZ Annotation}
\label{app:tikz-guidelines}

Human annotators follow a common checklist when converting mathematical figures into structured symbolic representations. The checklist emphasizes identification of core visual elements and faithful, compilable \LaTeX{}/TikZ authoring.

\begin{tcolorbox}[
    colback=gray!5,
    colframe=gray!45,
    title={Manual Image-to-TikZ Annotation Checklist},
    fonttitle=\bfseries,
    boxrule=0.6pt,
    arc=2mm,
    left=2mm,
    right=2mm,
    top=1mm,
    bottom=1mm
]
Inspect the source problem image and record the core visual information in the mathematical and geometric domains, including the coordinate-system type, function expressions, graphical boundaries, coordinates of special points, auxiliary lines, filled regions, and legend descriptions. Then manually author clear academic-style \LaTeX{} drawing code.

The requirements are as follows:
\begin{enumerate}
    \item Prefer the TikZ package to ensure the clarity of vector graphics.
    \item Strictly follow the parameter specifications in the original figure and do not omit any important details.
    \item Keep the code structure clean and well organized, with detailed comments to facilitate subsequent parameter adjustment.
    \item Provide a brief explanation indicating how the key parts of the code correspond to the visual elements in the original figure.
\end{enumerate}
\end{tcolorbox}

\section{Generated Token Analysis}
\label{app:generated-token-analysis}
% \begin{figure}[t] % [htbp] 指定图片的位置参数
%     \centering
%     \includegraphics[width=\linewidth]{fig/pdf/generate token.pdf} % 图片路径
%     \caption{Average generated token count across difficulty levels and input formats.} % 示例图的标题
%     \label{fig:g_token} % 图片标签，用于引用
% \end{figure}

\begin{figure*}[!t]
    \centering

    \begin{subfigure}{0.48\textwidth}
        \centering
        \includegraphics[width=\linewidth]{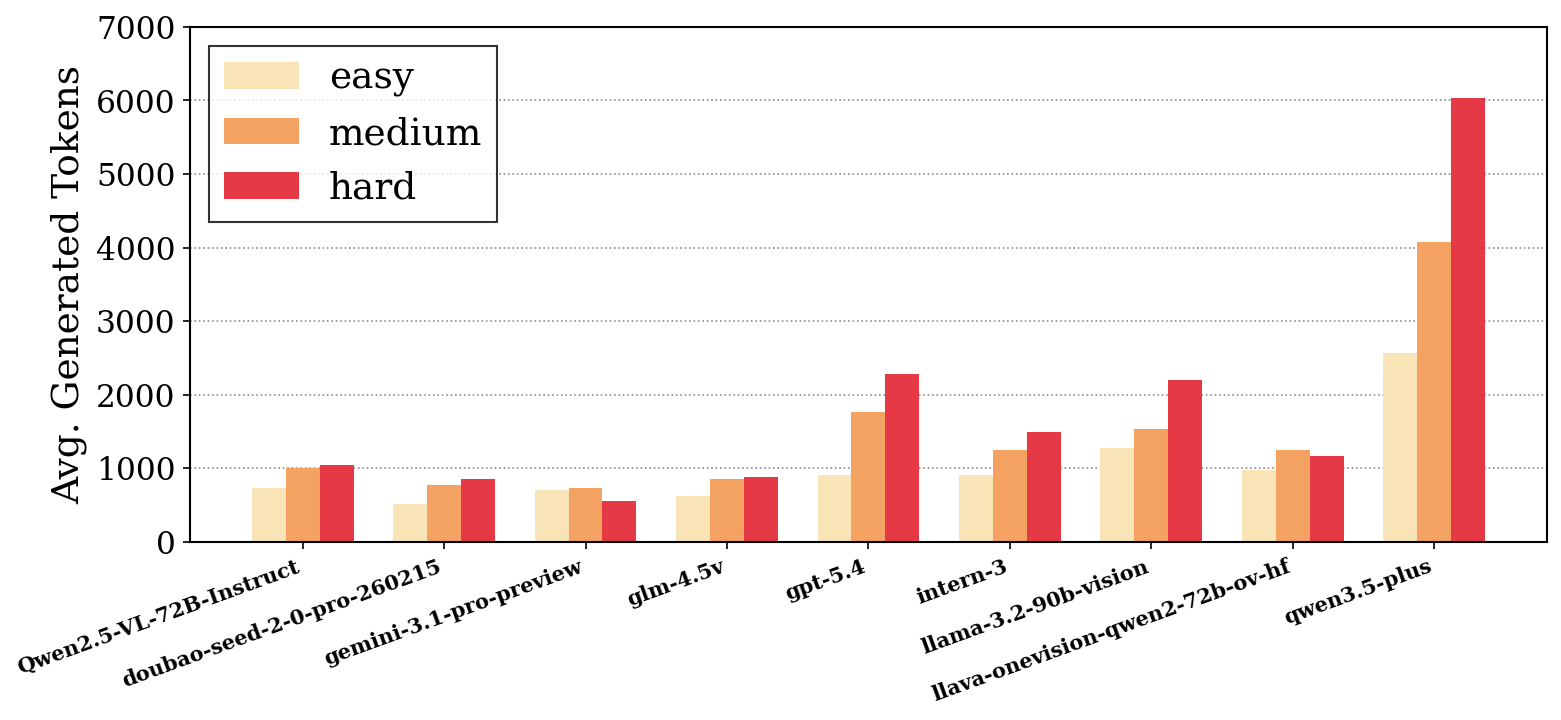}
        \caption{Only Text}
        \label{fig:tokens_only_text}
    \end{subfigure}
    \hfill
    \begin{subfigure}{0.48\textwidth}
        \centering
        \includegraphics[width=\linewidth]{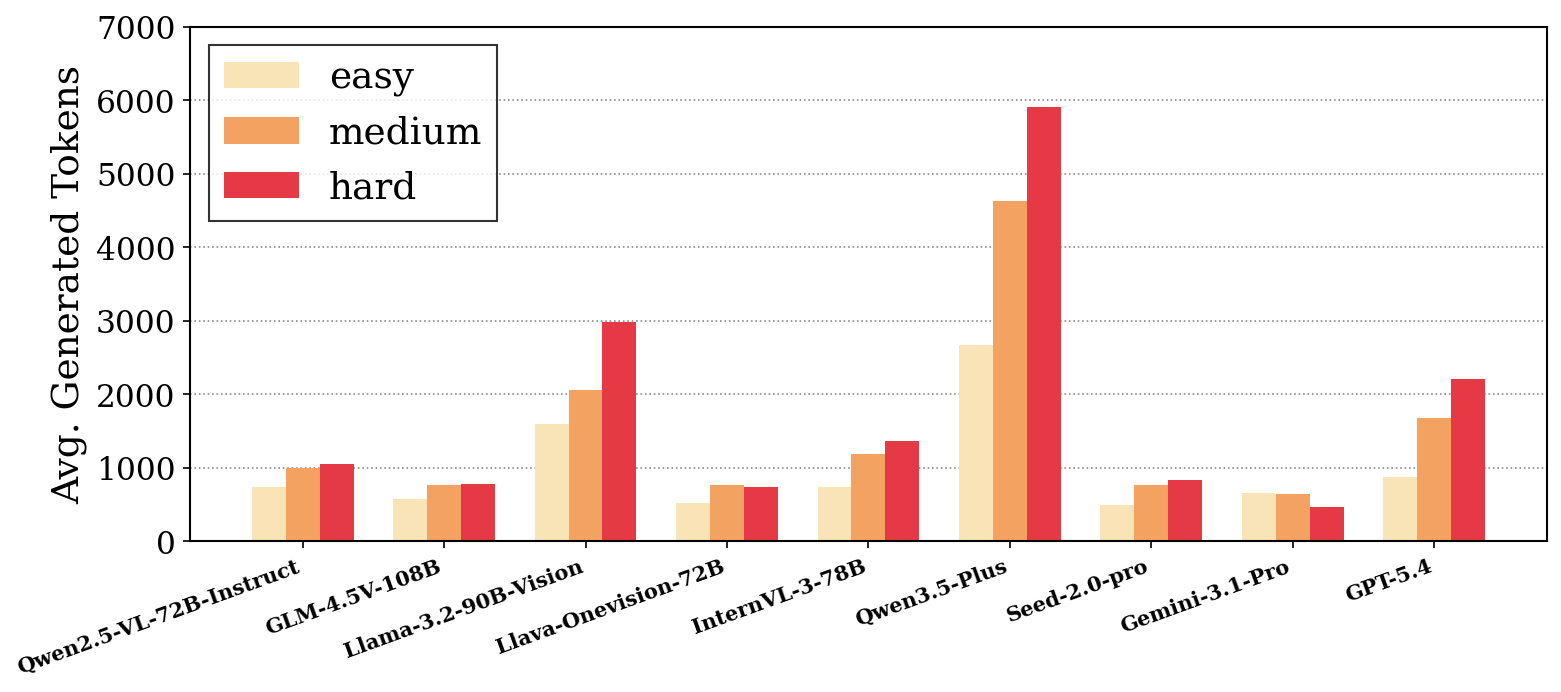}
        \caption{Text+Figure}
        \label{fig:tokens_with_image}
    \end{subfigure}

    \vspace{0.5em}

    \begin{subfigure}{0.48\textwidth}
        \centering
        \includegraphics[width=\linewidth]{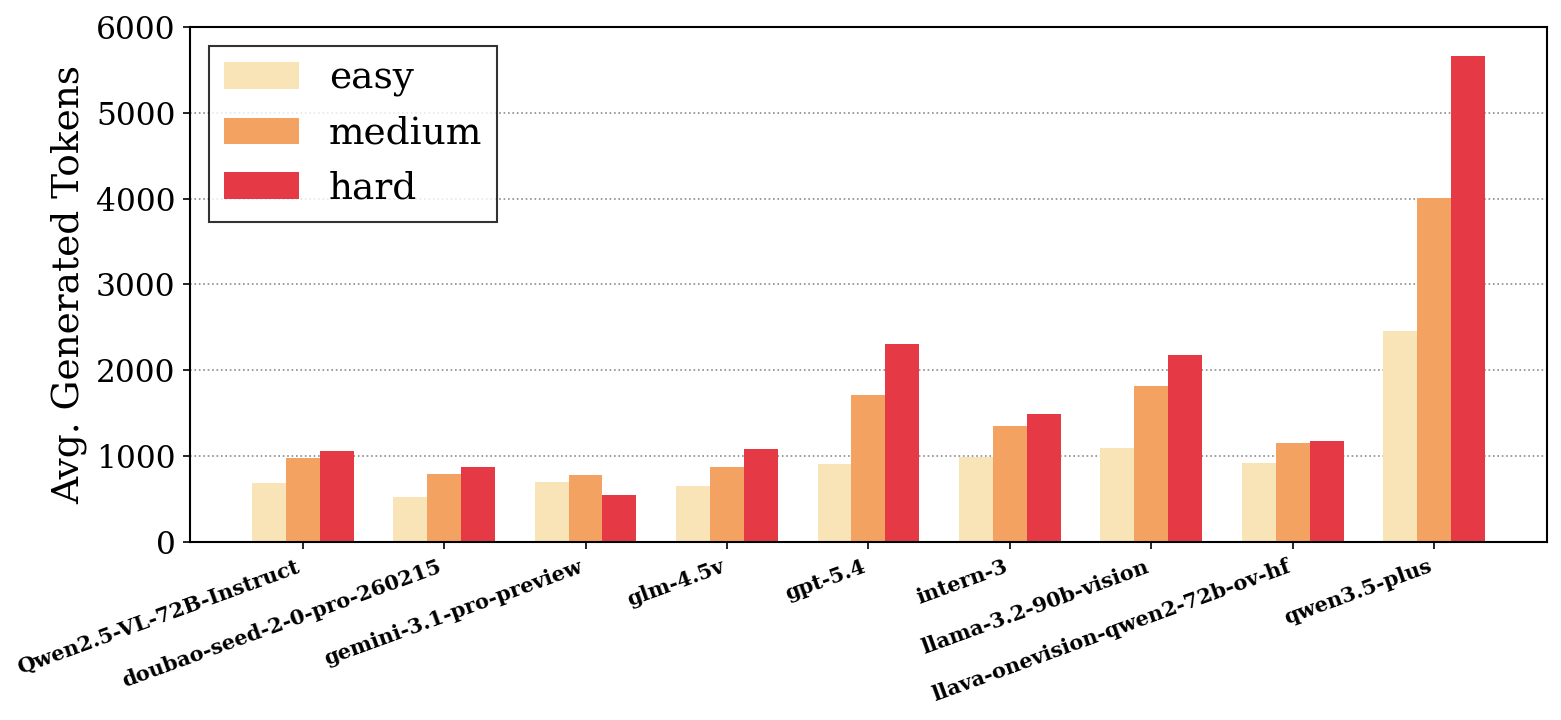}
        \caption{Text+\texttt{TikZ}}
        \label{fig:tokens_with_tikz}
    \end{subfigure}
    \hfill
    \begin{subfigure}{0.48\textwidth}
        \centering
        \includegraphics[width=\linewidth]{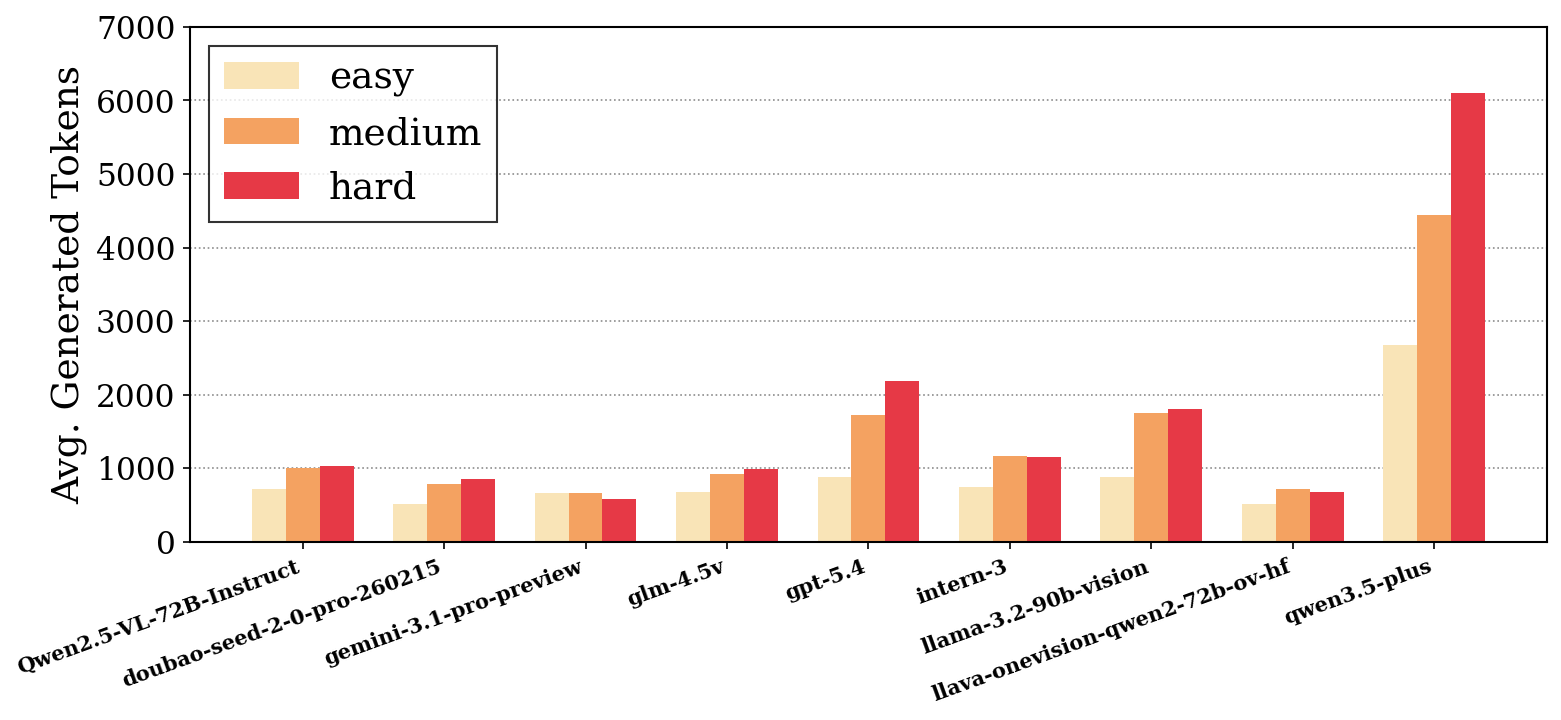}
        \caption{Text+Figure+\texttt{TikZ}}
        \label{fig:tokens_all}
    \end{subfigure}

    \caption{
    Average generated token count across difficulty levels and input formats.
    }
    \label{fig:g_token}
\end{figure*}

Figure~\ref{fig:g_token} reports the average number of generated tokens across difficulty levels and input formats.
The number of output tokens generally increases with problem difficulty, indicating that harder geometric problems tend to require longer reasoning chains.
In contrast, introducing additional input modalities, such as figures or \texttt{TikZ}, does not substantially increase the number of generated tokens.
This suggests that the performance changes observed in the main experiments are mainly associated with differences in available geometric information and multimodal fusion ability, rather than with a systematic increase in generation length or reasoning cost.

\section{Prompt for LLM-as-Judge Evaluation}
\label{app:llm-judge-prompt}

To evaluate the correctness of model-generated answers, we adopt an LLM-as-Judge evaluation protocol. 
Given a problem statement, the reference solution, and the student's answer, the judge model is instructed to assign a score based on answer-level correctness. 
For a problem requiring multiple answers, the maximum score is defined as the number of required answers, and each correctly answered item receives one point. 
An English translation of the Chinese evaluation prompt is shown below.

\begin{tcolorbox}[
    colback=gray!5,
    colframe=gray!45,
    title={Prompt Template for LLM-as-Judge Evaluation},
    fonttitle=\bfseries,
    boxrule=0.6pt,
    arc=2mm,
    left=2mm,
    right=2mm,
    top=1mm,
    bottom=1mm
]
You are a rigorous mathematics and geometry grading teacher. Please judge the student's solution according to the reference answer.

The grading rules are as follows:
\begin{enumerate}
    \item First determine the total score of the problem. If the problem requires multiple answers, the maximum score is equal to the number of required answers.
    \item Each correctly answered item receives 1 point, while incorrect items receive 0 points.
\end{enumerate}

Please output the result in JSON format, including the following fields:
\texttt{score} as an integer indicating the final score obtained by the student,
\texttt{max\_score} as an integer indicating the maximum score of the problem,
and \texttt{comment} as a brief explanation.

\medskip
\noindent\textbf{Problem:} \{question\}

\noindent\textbf{Reference Answer:} \{solution\}

\noindent\textbf{Student Answer:} \{answer\}
\end{tcolorbox}

\subsection{Reliability Analysis of LLM-based Evaluation}
\label{app:llm-judge-reliability}

\begin{figure}[t]
    \centering
    \begin{subfigure}{0.48\linewidth}
        \centering
        \includegraphics[width=\linewidth]{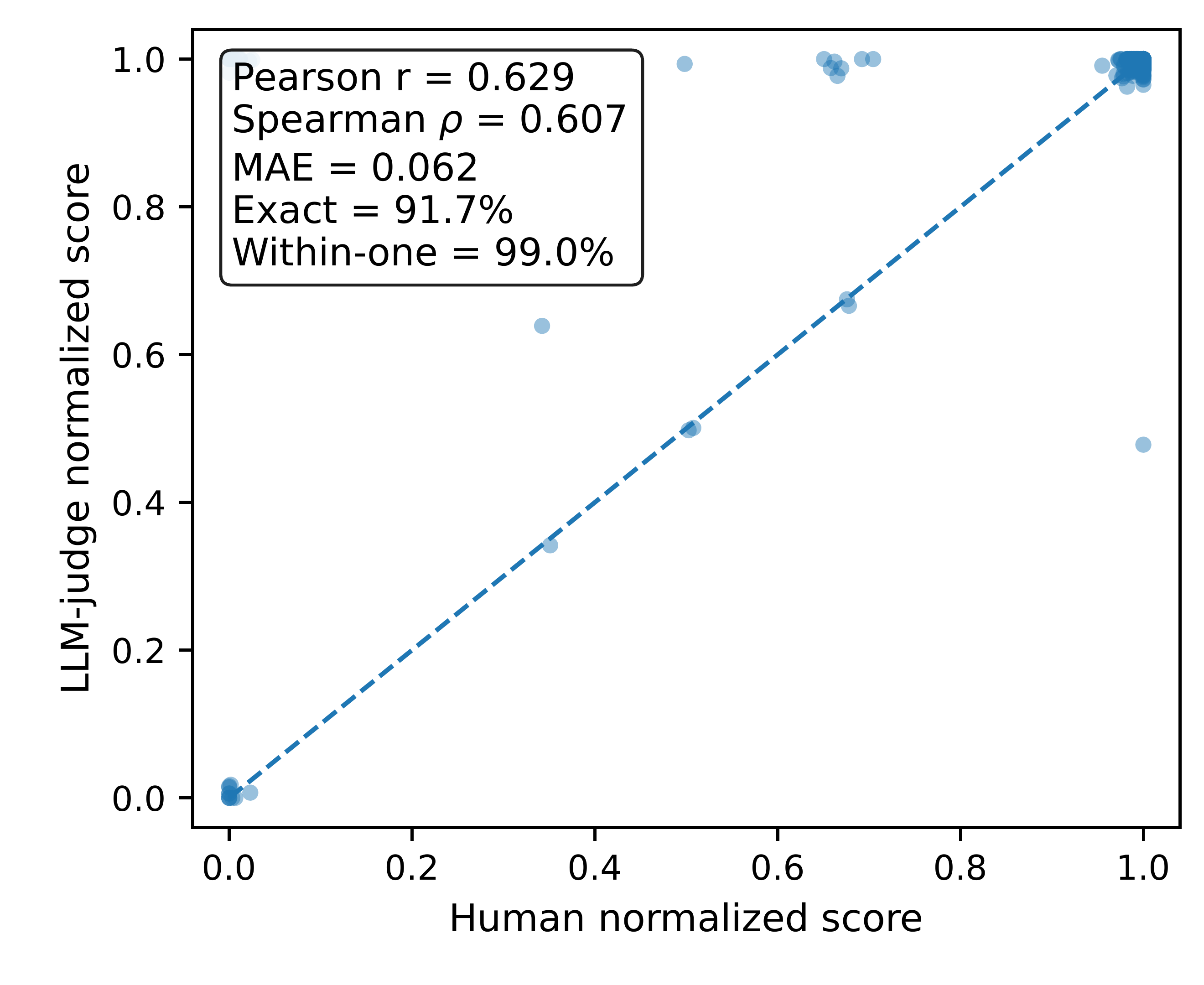}
        \caption{Score-level agreement}
        \label{fig:judge_score_agreement}
    \end{subfigure}
    \hfill
    \begin{subfigure}{0.48\linewidth}
        \centering
        \includegraphics[width=\linewidth]{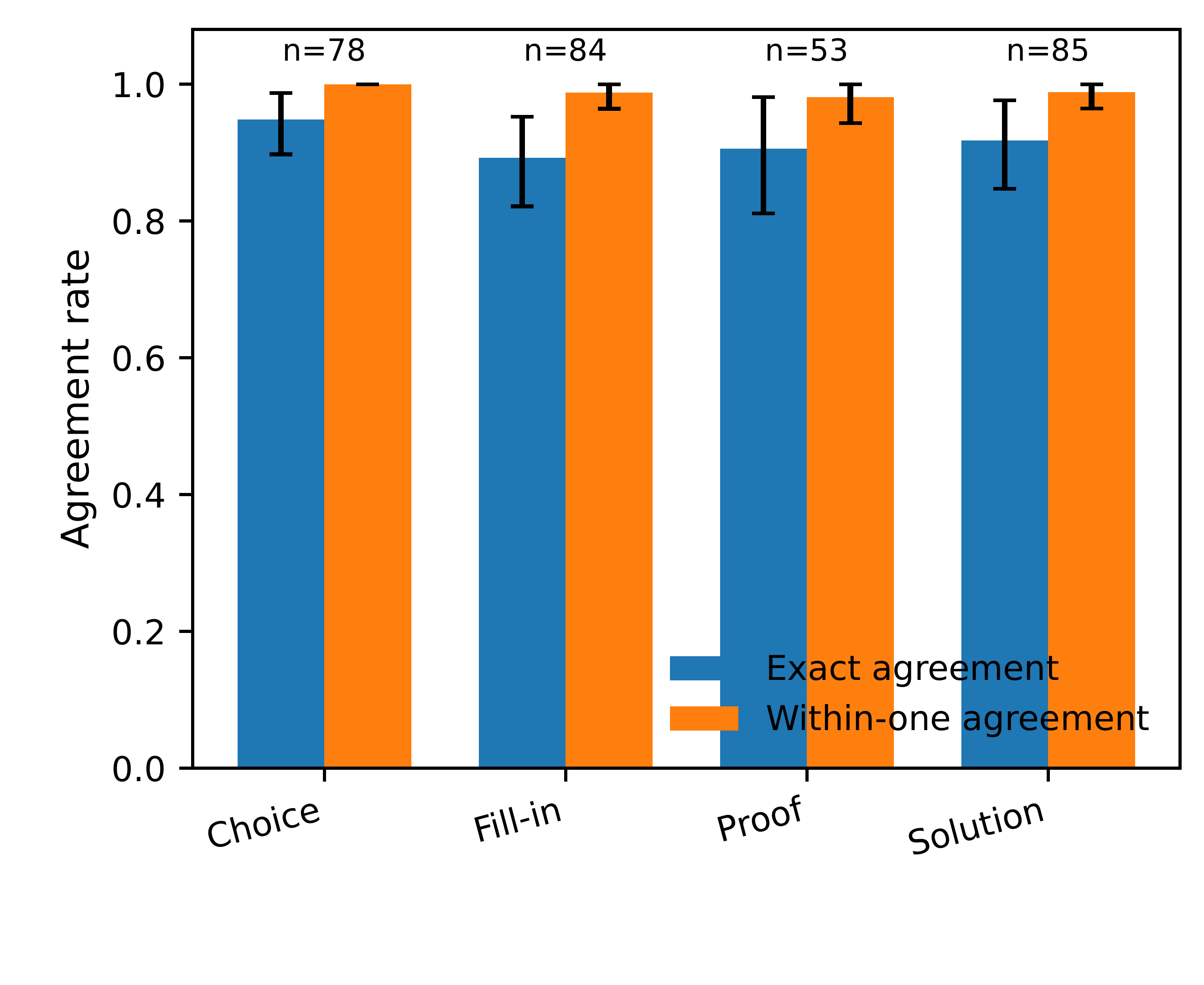}
        \caption{Agreement by question type}
        \label{fig:judge_type_agreement}
    \end{subfigure}

    \caption{
    Agreement between the LLM-based evaluator and human audit results.
    Left: normalized score-level correlation between human scores and LLM-judge scores.
    Right: exact agreement and within-one-point agreement across question types, with 95\% bootstrap confidence intervals.
    }
    \label{fig:llm_judge_reliability}
\end{figure}

To assess the LLM-based evaluator, we conduct a human-audit agreement analysis on 300 sampled responses: 78 choice, 84 fill-in, 53 proof, and 85 solution items. Human annotators independently score each response using the same criteria as the LLM judge, and we compare the scores after normalization by each problem's maximum score.

Figure~\ref{fig:judge_score_agreement} reports Pearson $r=0.629$, Spearman $\rho=0.607$, and normalized-score MAE $=0.062$. Exact agreement is 91.7\% and within-one-point agreement is 99.0\%. Figure~\ref{fig:judge_type_agreement} further breaks down agreement by question type. The point estimates are higher for objective formats than for proof- and solution-style problems, where valid reasoning paths are more diverse.

Overall, this audit supports the fixed grader for large-scale descriptive evaluation while motivating explicit JSON scoring rules for open-ended responses.

\section{Translations of the Dataset Examples}
\label{app:translations}

Figure~\ref{fig:geo_example} preserves the Chinese excerpts in the dataset examples. The translations and toneless Pinyin romanizations below follow the visible excerpts, including their ellipses; they do not reconstruct conditions or derivations omitted from the figure. Mathematical symbols retain their original notation.

\subsection{Easy Example}

\textbf{Question (English).} In triangle $ABC$, angle $A$ is $60^\circ$, side $AB$ is $2$, and side $BC$ is $\sqrt{6}$. The angle bisector of $A$ meets $BC$ at $D$. Find the length of $AD$.

\textbf{Question (Pinyin).} Zai sanjiaoxing $ABC$ zhong, yizhi jiao $A$ wei $60$ du, bian $AB$ wei $2$, bian $BC$ wei $\sqrt{6}$, jiao $A$ de jiao pingfenxian jiao $BC$ yu dian $D$, yaoqiu jisuan $AD$ de changdu.

\textbf{Answer excerpt (English).} Answer: $2$. Detailed solution: As shown, let $AB=c$, $AC=b$, and $BC=a$. \ldots{} Therefore, $C=45^\circ$ and $B=180^\circ-60^\circ-45^\circ=75^\circ$.

\textbf{Answer excerpt (Pinyin).} Da'an: $2$. Xiangjie: Ru tu suo shi, ji $AB=c$, $AC=b$, $BC=a$. \ldots{} Suoyi $C=45^\circ$, $B=180^\circ-60^\circ-45^\circ=75^\circ$.

\subsection{Medium Example}

\textbf{Question excerpt (English).} Let $f(x)$ be a cubic function \ldots{} Its local maximum and minimum values are, respectively: A.\ $f(1)$ and $f(-1)$; \ldots{} D.\ $f(-2)$ and $f(2)$.

\textbf{Question excerpt (Pinyin).} She $f(x)$ shi sanci hanshu \ldots{}, jida zhi yu jixiao zhi fenbie shi: A.\ $f(1)$ yu $f(-1)$; \ldots{} D.\ $f(-2)$ yu $f(2)$.

\textbf{Answer excerpt (English).} Answer: D. Detailed solution: When $x<0$, $f'(x)$ and $x f'(x)$ have opposite signs; \ldots{} when $x=2$, $f(x)$ has a local minimum $f(2)$.

\textbf{Answer excerpt (Pinyin).} Da'an: D. Xiangjie: Dang $x<0$ shi, $f'(x)$ de fuhao yu $x f'(x)$ de fuhao xiangfan; \ldots{} dang $x=2$ shi, $f(x)$ you jixiao zhi $f(2)$.

\subsection{Hard Example}

\textbf{Question excerpt (English).} Located in Yangpu \ldots{} the upper and lower bases \ldots{} (1) Prove: \ldots{} $BC\perp C_1F$. (2) Find the distance from point $O$ to the plane \ldots{}.

\textbf{Question excerpt (Pinyin).} Zuoluo yu Yangpu \ldots{} de shangxia liang dimian de. (1) Qiuzheng: \ldots{} $BC\perp C_1F$. (2) Qiu dian $O$ dao ping \ldots{} juli.

\textbf{Answer excerpt (English).} Detailed solution: (1) Proof: Connect $CF$. Since the base is a regular octagon \ldots{}, $BC\perp C_1F$. (2) Connect $OC_1$ and $OG_1$ \ldots{} the distance to $ACG_1$ is $32$.

\textbf{Answer excerpt (Pinyin).} Xiangjie: (1) Zhengming: Lianjie $CF$, yinwei dimian wei zheng bajiaoxing \ldots{}, suoyi $BC\perp C_1F$. (2) Lianjie $OC_1$, $OG_1$ \ldots{} $ACG_1$ de juli wei $32$.

\subsection{Annotation Interface Labels}
\label{app:interface-translations}

The screenshots in Figure~\ref{fig:annotation-system} illustrate the interface rather than additional evaluation examples. The following glossary translates the Chinese controls and headings (toneless Pinyin in italics).

\begin{itemize}
    \item \emph{Yuan tu zhanshi} / \emph{yuan ti tupian}: source image; \emph{dianji tupian ke fangda chakan}: click the image to enlarge.
    \item \emph{Rengong biaozhu quyu}: manual annotation area; \emph{timu} / \emph{wenti}: question; \emph{jieda}: solution; \emph{da'an}: answer; \emph{fenxi}: analysis; \emph{xiangjie}: detailed solution.
    \item \emph{Tupian biaozhu} / \emph{tuxing biaozhu}: image/figure annotation; \emph{biaozhu tupian}: annotated image; \emph{wei wancheng}: incomplete.
    \item \emph{Mei tiao baohan}: each entry contains; \emph{AI biaozhu jieguo}: AI annotation result; \emph{rengong biaozhu jieguo}: human annotation result; \emph{duiying jietu}: corresponding screenshot.
    \item \emph{Ke kong} / \emph{kexuan}: optional; \emph{bitian}: required field; \emph{bichuan}: required upload; \emph{weizhi}: position; \emph{timu bufen}: question section.
    \item \emph{Tishi: dianji shanchu anniu ke yichu tupian}: tip: click delete to remove an image; \emph{zanwu AI biaozhu jieguo}: no AI annotation result yet.
    \item \emph{Ni zuizhong queren de TikZ / LaTeX}: your final confirmed TikZ / LaTeX; \emph{dianji cichu hou Ctrl+V zhantie jietu}: click here and press Ctrl+V to paste a screenshot.
    \item \emph{Huo shangchuan tupian (beiyong)}: or upload an image (alternative); \emph{dianji shangchuan}: click to upload; \emph{shiyong da moxing shengcheng}: generate using a large model; \emph{tianjia biaozhu tupian}: add an annotated image.
    \item \emph{Shenhe juece}: review decision; \emph{dangqian zhuangtai} / \emph{zhuangtai}: current status / status; \emph{shenhe yijian}: review comments; \emph{tongguo}: accept; \emph{bohui}: reject.
    \item \emph{Jietu}: screenshot; \emph{LaTeX bianyi jieguo}: LaTeX compilation result; \emph{tuxing timu-1} / \emph{tuxing da'an-1}: question figure 1 / answer figure 1.
\end{itemize}

\paragraph{Circle problem displayed in the review interface.}
\textbf{English.} As shown, $AB$ is a diameter of circle $O$, and $AD$ and $BC$ are tangents. $CO$ bisects $\angle BCD$ and intersects the circle at $E$; connect $BE$. (1) Prove that $CD$ is tangent to circle $O$. (2) If $AD=1$ and $CD=4$, find $BE$. Answer: (1) see the solution; (2) $\sqrt{3}$. The visible solution draws $OF\perp CD$ at $F$. Since $BC$ is tangent, $OB\perp BC$. The angle bisector gives $OF=OB$, so $OF$ is a radius and $CD$ is tangent. Tangent lengths give $AD=DF$, $CF=BC$, $CD=AD+BC$, and $BC=3$.

\textbf{Pinyin.} Ru tu, $AB$ shi yuan $O$ de zhijing, $AD$ he $BC$ fenbie shi yuan $O$ de qiexian, $CO$ pingfen $\angle BCD$, qie yu yuan $O$ jiao yu dian $E$, lianjie $BE$. (1) Qiuzheng: $CD$ shi yuan $O$ de qiexian. (2) Ruo $AD=1$, $CD=4$, qiu $BE$ de chang. Da'an: (1) jian jiexi; (2) $\sqrt{3}$. Xiangjie: guo dian $O$ zuo $OF\perp CD$ yu $F$. Yinwei $BC$ shi yuan $O$ de qiexian, suoyi $OB\perp BC$ yu $B$. You yinwei $CO$ pingfen $\angle BCD$, suoyi $OF=OB$. $OB$ shi yuan $O$ de banjing, suoyi $OF$ ye shi yuan $O$ de banjing, $CD$ shi yuan $O$ de qiexian. You (1) de $CD$ shi yuan $O$ de qiexian, qiedian wei $F$; $AD$ he $BC$ fenbie shi yuan $O$ de qiexian; $AD=DF$, $CF=BC$, $CD=AD+BC$, $BC=4-1=3$.

\end{document}